\documentclass[10pt,twocolumn,letterpaper]{article}

\usepackage[pagenumbers]{cvpr} 
\usepackage{multirow}
\usepackage{soul}
\usepackage{amssymb}
\usepackage{amsmath}
\definecolor{cvprblue}{rgb}{0.21,0.49,0.74}
\usepackage[pagebackref,breaklinks,colorlinks,allcolors=cvprblue]{hyperref}

\def\modelname{{FAM diffusion}} 

\usepackage{colortbl}

\definecolor{cyan}{cmyk}{.3,0,0,0}
\definecolor{hr}{gray}{0.5}
\definecolor{hg}{gray}{0.8}
\def\onedot{.}

\definecolor{cyan}{cmyk}{.3,0,0,0}
\definecolor{hr}{gray}{0.5}

\newlength\savewidth

\usepackage{subcaption} 

\def\paperID{15603} 
\def\confName{CVPR}
\def\confYear{2025}

\title{FAM Diffusion: Frequency and Attention Modulation for High-Resolution Image Generation with Stable Diffusion}

\author{%
  Haosen Yang\textsuperscript{1,2}\thanks{This work was conducted while Haosen Yang was an intern at Samsung AI Center, Cambridge, UK.} \quad
  Adrian Bulat\textsuperscript{1} \quad
  Isma Hadji\textsuperscript{1} \quad
  Hai X. Pham\textsuperscript{1} \quad
  Xiatian Zhu\textsuperscript{2} \quad \\
  Georgios Tzimiropoulos\textsuperscript{1,3} \quad
  Brais Martinez\textsuperscript{1} \\
  \\
  \textsuperscript{1}Samsung AI Center, Cambridge, UK \quad
  \textsuperscript{2}University of Surrey, UK \quad 
  \textsuperscript{3}Queen Mary University, UK\\
}

\begin{document}
\maketitle
\begin{abstract}
Diffusion models are proficient at generating high-quality images. They are however effective only when operating at the resolution used during training. Inference at a scaled resolution leads to repetitive patterns and structural distortions. 
Retraining at higher resolutions quickly becomes prohibitive. Thus, methods enabling pre-existing diffusion models to operate at flexible test-time resolutions are highly desirable.
Previous works suffer from frequent artifacts and often introduce large latency overheads. 
We propose two simple modules that combine to solve these issues. We introduce a Frequency Modulation (FM) module that leverages the Fourier domain to improve the global structure consistency, and an Attention Modulation (AM) module which improves the consistency of local texture patterns, a problem largely ignored in prior works.
Our method, coined \modelname{}, can seamlessly integrate into any latent diffusion model and requires no additional training. 
Extensive qualitative results highlight the effectiveness of our method in addressing structural and local artifacts, while quantitative results show state-of-the-art performance. Also, our method avoids redundant inference tricks for improved consistency such as patch-based or progressive generation, leading to negligible latency overheads.
\end{abstract}

\section{Introduction}

\begin{figure*}[t]
    \centering
    \subfloat[Direct Inference]{
        \includegraphics[width=0.24\linewidth]{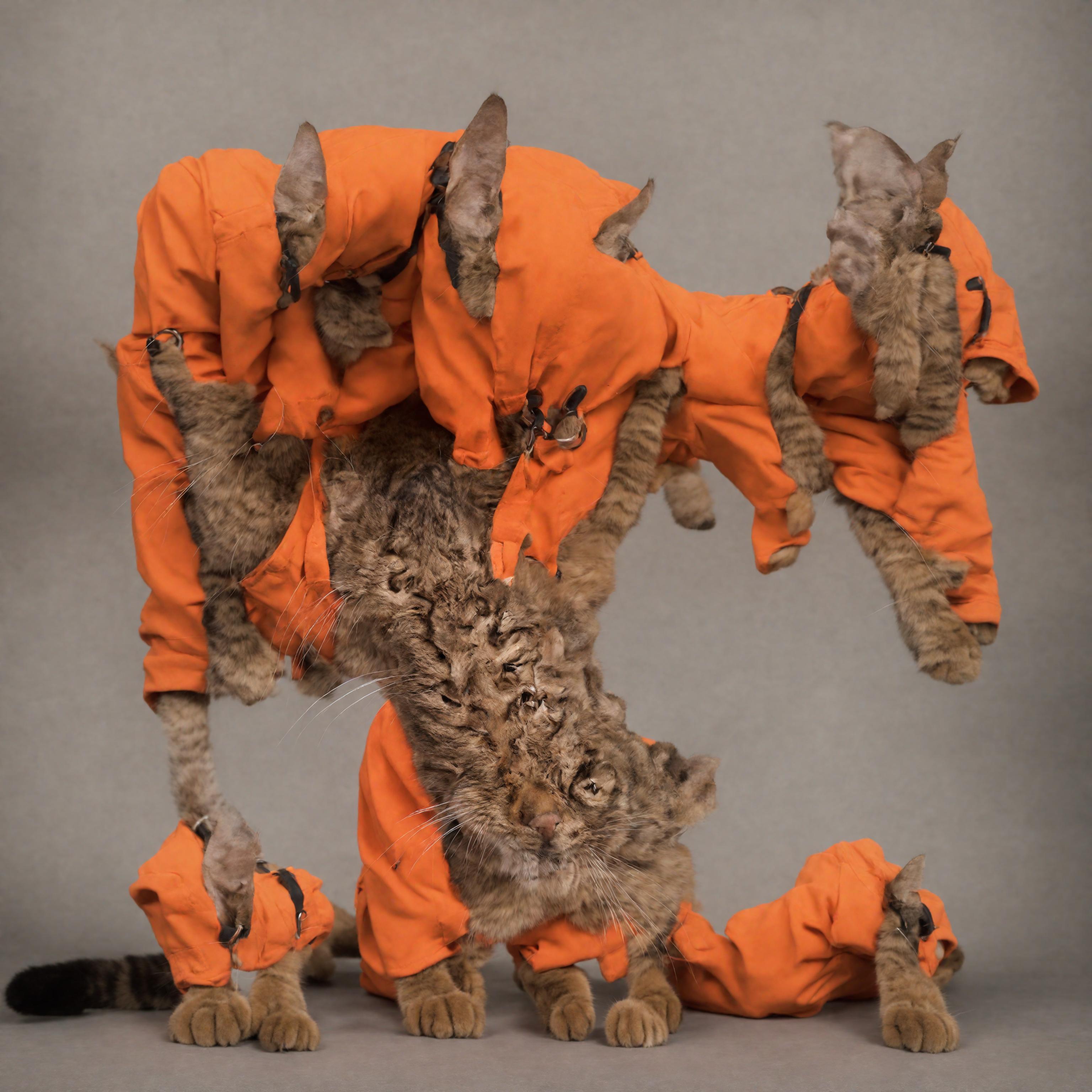}
    }
    \subfloat[DemoFusion]{
        \includegraphics[width=0.24\linewidth]{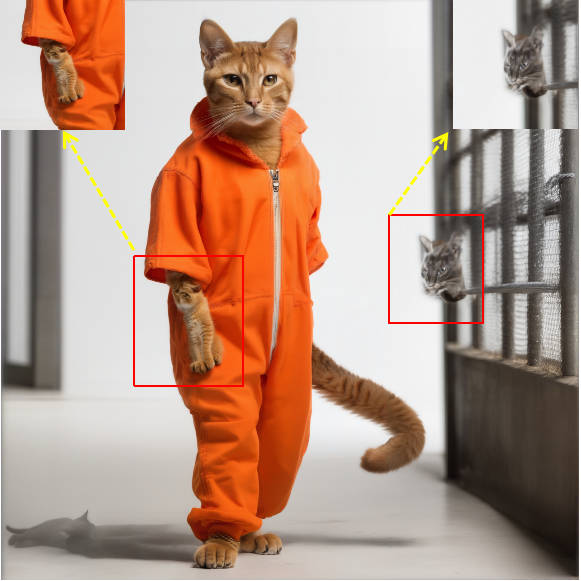}
    }
    \subfloat[HiDiffusion]{
        \includegraphics[width=0.24\linewidth]{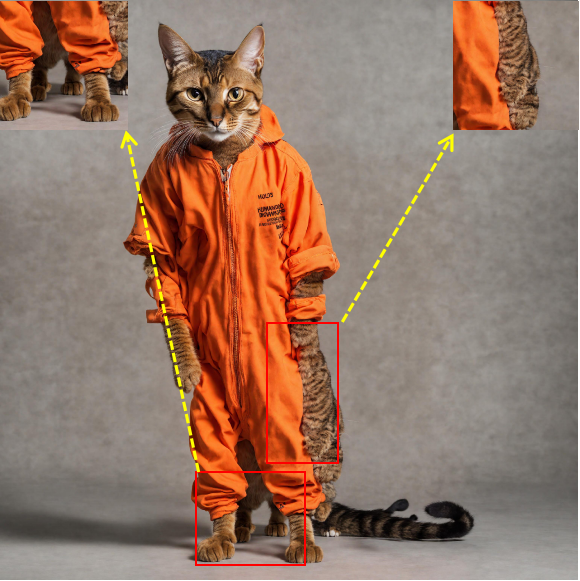}
    }
    \subfloat[Ours]{
        \includegraphics[width=0.24\linewidth]{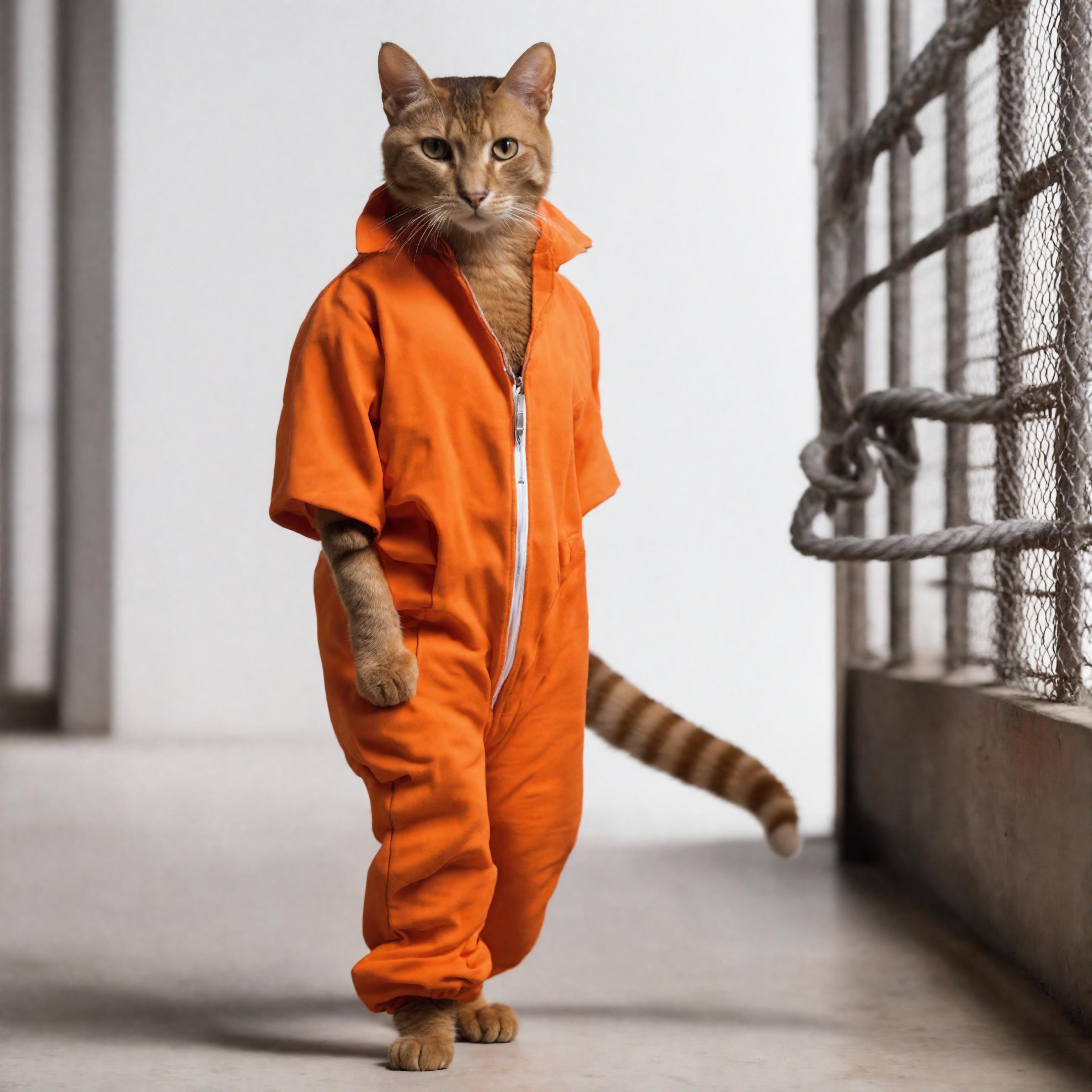}
    }
    \caption{Comparisons of 3× (3072 × 3072) image generation based on SDXL~\cite{sdxl}.}
    \label{fig:fig1}
\end{figure*}

Diffusion models \cite{ldm_cvpr22} demonstrate impressive generative power across a range of applications \cite{wang2024single, ruiz2023dreambooth, zhang2023adding, noroozi2024you, poole2022dreamfusion, he2022latent,wu2023tune}. While powerful, one known shortcoming of diffusion models is their inability to seamlessly scale to higher resolutions beyond the one used during training. It is known that directly generating images at resolutions beyond the training resolution results in severe object repetition and unrealistic local patterns~\cite{multidiffusion, scalecrafter, demofusion}. This is illustrated in Figure~\ref{fig:fig1}(a). While retraining diffusion models on higher-resolution images is a straightforward solution, the computational demands quickly become prohibitive. This restricts applications requiring flexible or high-resolution image generation, e.g. 4K. Therefore, adapting pre-trained diffusion models to generate high-resolution images without additional training is a topic of high interest that we tackle in this work.

Prior efforts addressing this important problem can be largely categorized into two tracks. The first set of approaches, e.g. \cite{demofusion,accdiffusion}, propose mechanisms that improve the global structure consistency by steering the high-resolution generation using the image generated at native (i.e. training) resolution. However, the effectiveness of such mechanisms is mixed, with trailing issues like poor detail quality, inconsistent local textures, and even persisting pattern repetitions as shown in Figure~\ref{fig:fig1}(b). Furthermore, these works typically operate on a patch-based basis, generating one patch at a time. Concretely, this means that these methods resort to redundant and overlapping forward passes, leading to large latency overheads.
The second group of approaches, e.g. \cite{scalecrafter, fouriscale, hidiffusion}, eschews patch-based generation in favor of a one-pass approach by directly altering the model architecture. This leads to faster generation, but unfortunately, it comes at the cost of image quality, as shown in Fig.~\ref{fig:fig1} (c).

To address the aforementioned limitations, we propose a straightforward yet effective approach that takes the best of both worlds. 
Our method follows the single pass generation strategy for improved latency but, like patch-based approaches, leverages the native resolution generation to steer the high-resolution one.
Specifically, our method starts by generating an image at native resolution conditioned on the input text prompt. We then resort to a test-time diffuse-denoise strategy~\cite{song2020denoising, demofusion, hertz2022prompt}, where the high-resolution denoising stage is guided by the native resolution diffusion process. However, instead of blindly steering the high-res image toward the low-res one as done elsewhere \cite{demofusion,accdiffusion}, we propose a Frequency Modulation (FM) module. In particular, we leverage the Fourier domain to selectively condition low-frequency components during the high-resolution image generation stage, while providing full control over high-frequency components to the denoising process. 

While the FM module resolves artifacts related to global consistency, artifacts related to inconsistent local texture might still be present, i.e. finer texture generated on semantically related parts of the image might be inconsistent.
To tackle this second issue, largely ignored in the literature, we propose an Attention Modulation (AM) mechanism that leverages attention maps from the denoising process at native resolution to condition the attention maps of the denoising process at high resolution. Since attention maps at native resolution encode which regions of the image are semantically related, they regularize the high-res denoising towards consistent finer texture generation. Our method, coined \textbf{F}requency and \textbf{A}ttention \textbf{M}odulated diffusion (\modelname{}), combines our FM and AM modules to yield superior quality results, see Fig.~\ref{fig:fig1} (d).

Our method seamlessly integrates with any latent diffusion model without additional training or architectural changes. We empirically show that our method significantly enhances the quality and efficiency of high-resolution image generation, establishing a new state-of-the-art.

\section{Related Work}

Diffusion models have shown impressive performance in generating creative and accurate representations given text prompts \cite{ddpm, ldm_cvpr22}. While early work \cite{ldm_cvpr22} was limited to generating relatively low-resolution images (i.e. $256\times 256$), follow-up work showed that their performance can scale to higher resolutions, e.g. $512\times512$ with SD1.5 \cite{ldm_cvpr22} and $1024\times1024$ with SDXL \cite{sdxl}. However, a major shortcoming with all these models is that generation remains limited by the resolution used at training time. Naively targeting higher train-time resolutions quickly results in prohibitive training costs and computational requirements, and the limited availability of high-resolution training data also restricts the diversity of image generation. Thus, adapting pre-trained diffusion models to generate high-resolution images without retraining has emerged as a topic of interest.

Early works \cite{multidiffusion,syncdiffusion} proposed using overlapping patches at native resolution and blending the outputs to produce an image without seams. However, this leads to frequent repetitions and inconsistent global image structure. Therefore, subsequent works introduced various mechanisms to encourage global structural consistency. For instance, DemoFusion \cite{demofusion} proposed a patch-based generation process with mechanisms such as skip residuals and progressive upsampling, while AccDiffusion \cite{accdiffusion} used localized prompting to guide high-resolution generation and improve consistency with images generated at native resolutions. 
However, these methods still suffer from issues like local repetitions, and inconsistent global coherence. They also have significant latency overheads due to 
the running cost of multiple backward passes.
To mitigate the high latencies, other works aim to generate high-resolution images in a single pass by modifying the architecture of the UNet. For example, ScaleCrafter~\cite{scalecrafter} employs dilated convolutions to adjust the receptive field of convolutions in the denoising UNet. HiDiffusion~\cite{hidiffusion} introduces an alternative UNet that dynamically adjusts the feature map size during the denoising process. While these approaches achieve faster generation, they often result in image distortions.

More closely related to ours are methods that have approached structural consistency from a frequency domain perspective. FouriScale~\cite{huang2024fouriscale} splits the image in Fourier domain, then proceeds to incorporate a low-pass filtering operation and impose structural consistency with an image generated at natire resolution. However, this splitting operation results in unrealistic images. HiPrompt~\cite{liu2024hiprompt} decomposes images into spatial frequency components conditioned on local and global prompts, but it often relies on redundant operations that lead to high latencies. ResMaster~\cite{shi2024resmaster} leverages low-frequency information from the latent representation of the native image to provide desirable global semantics during the denoising process. However, it ignores the noise distribution differences between the current high-resolution denoising step and the native image in latent space. In addition, it still relies on patch-based denoising, making it inefficient. In contrast to these methods, we propose a one-pass method that does not alter the model architecture. 
Importantly, our method introduces a complementary novel attention modulation mechanism, which targets local structure consistency; an issue overlooked by all existing works.

\section{Method}
\label{sec:method}

\begin{figure*}
    \centering
    \includegraphics[width=0.9\textwidth]{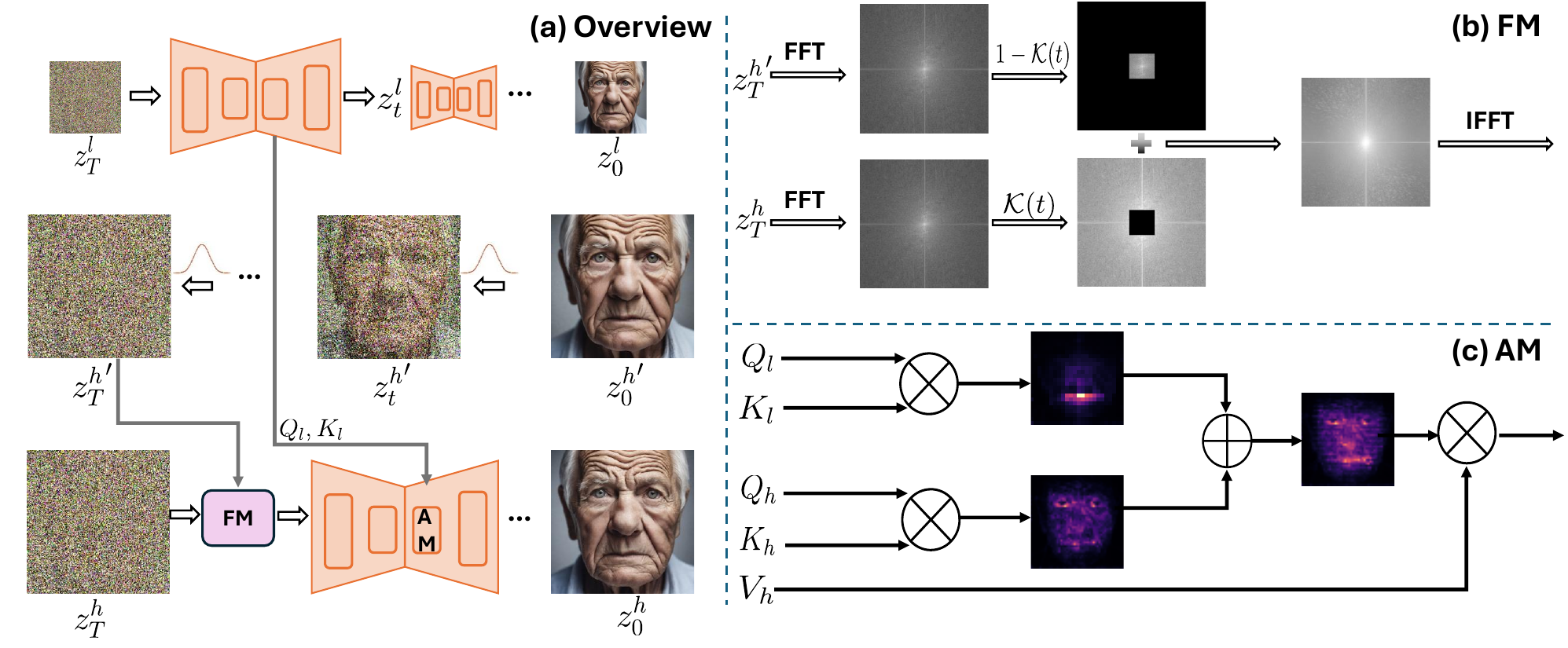}
    \caption{
Overview of the~\modelname{}. (a) We first generate an image at native resolution, followed by a test-time diffuse-denoise process. We incorporate our Frequency Modulation module and Attention Modulation during high-res denoising to control global structure and fine local texture, respectively.
(b) Details of the Frequency Modulation, where we use the Fourier domain to selectively condition low-frequency components during high-res denoising while leaving high-frequency components fully controllable.
(c) Details of Attention Modulation, where attention maps from the native image denoising are used to correct the high-res denoising.
}
    \label{fig:method}
\end{figure*}

\begin{figure*}[t] 
    \centering
    \subfloat[DI]{
        \includegraphics[width=0.18\linewidth]{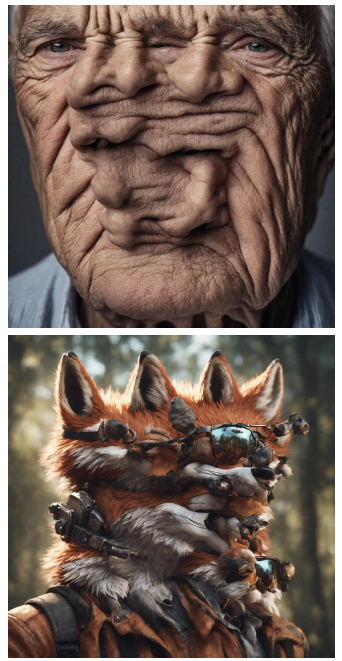}
        \label{fig:ablation_all_di}
    }
    \subfloat[DI*]{
        \includegraphics[width=0.18\linewidth]{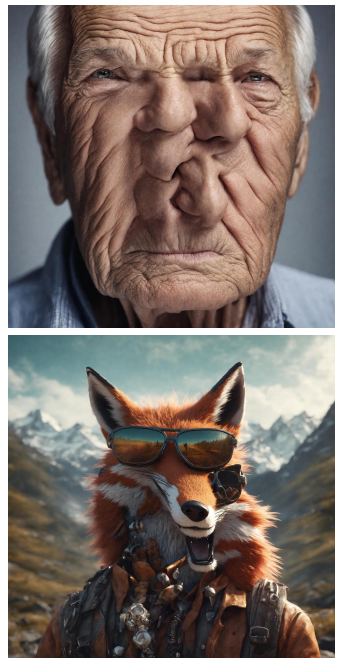}
        \label{fig:ablation_all_di_star}
    }
    \subfloat[SR]{
        \includegraphics[width=0.18\linewidth]{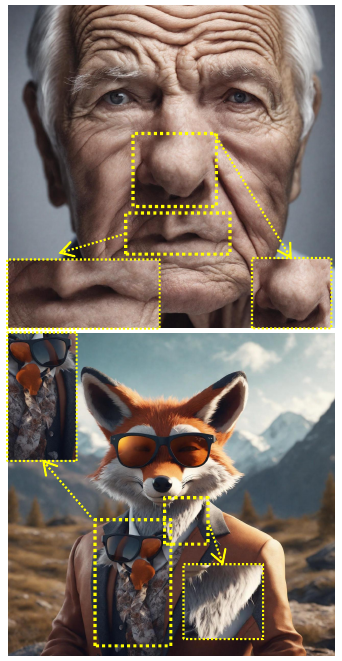}\label{fig:ablation_all_sr}
    }
    \subfloat[FM]{
        \includegraphics[width=0.18\linewidth]{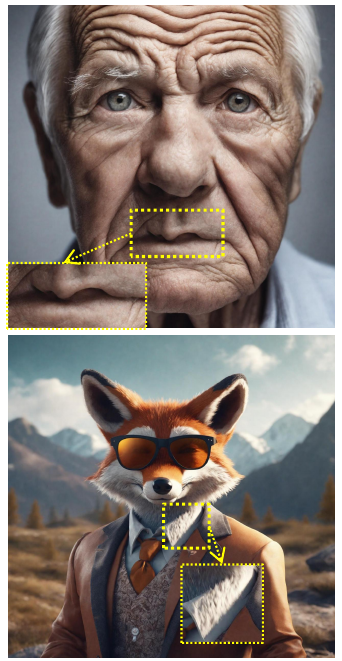}\label{fig:ablation_all_fm}
    }
    \subfloat[FM-AM ]{
        \includegraphics[width=0.18\linewidth]{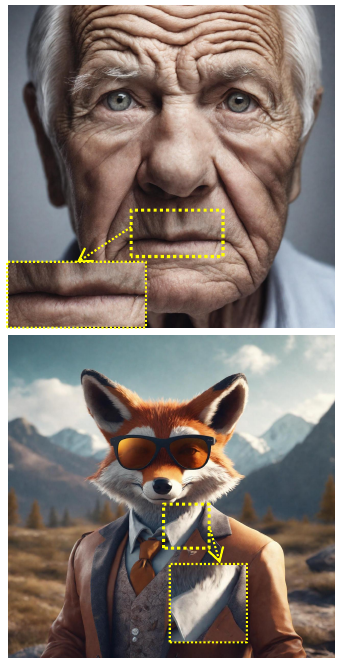}\label{fig:ablation_all_fm_am}
    }
    \caption{Ablation on the components of~\modelname{}. Direct Inference (DI) at high resolution from noise, Direct Inference from low-res latent (DI*), Skip Residual (SR) from DemoFusion~\cite{demofusion}, Frequency Modulation (FM), Attention Modulation (AM).}
    \label{fig:ablation_all}
\end{figure*}

In this work, we leverage pretrained latent diffusion models (LDMs), which have been extensively trained on large-scale high-quality data. Our goal is to generate images at higher resolutions than during training, without any additional finetuning or model modification. Sec.~\ref{ssec:preliminaries} briefly reviews the diffusion notation and the test-time diffuse-denoise strategy. In Sec.~\ref{ssec:feature-aiding} we present our \textit{Frequency Modulated} (FM) denoising approach, which is designed to improve global consistency. Finally, we introduce our Attention Modulation (AM) mechanism, which is designed to improve the consistency of the local texture and high-frequency detail, in Sec.~\ref{ssec:attention_mixing}. We provide an overview of our method in Figure \ref{fig:method}.

\subsection{Preliminaries}
\label{ssec:preliminaries}

\noindent \textbf{Latent Diffusion Models (LDM)~\cite{ldm_cvpr22}}: We operate in the realm of LDMs, which 
first convert image $\mathbf{x}_0$ to a latent representation $\mathbf{z}_0$ using an encoder such that $\mathbf{z}_0=\mathcal{E}(\mathbf{x}_0)$, $\mathbf{z}_0 \in \mathbb{R}^{c \times h \times w}$.
During training, a Markovian diffusion process progressively adds noise to the input latent $\mathbf{z}_0$ according to a predefined schedule $\beta_t, t \in [1,T]$ by sampling sequentially from: 
\begin{equation}
\label{eq:diffusion}
q(\mathbf{z}_t|\mathbf{z}_{t-1}):=\mathcal{N}( \mathbf{z}_t | \sqrt{1-\beta_t} \mathbf{z}_{t-1},  \beta_t \mathbf{I})
\end{equation}
Conversely, a trainable denoising process progressively recovers the original latent $\mathbf{z}_0$ using a noise estimator $\mathcal{Z}_\theta=\left(\mu_{\theta},\Sigma_{\theta}\right)$ parametrized by $\theta$ by sampling from:

\begin{equation}
\label{eq:denoise}
p_{\theta}(\mathbf{z}_{t-1}|\mathbf{z}_t):=\mathcal{N}\left(\textbf{z}_{t-1} | \mu_{\theta}(\mathbf{z}_t, t), \Sigma_{\theta}(\mathbf{z}_t, t) \right)
\end{equation}
During inference, an image is generated by denoising from random noise, $\mathbf{z_T} \sim \mathcal{N}(0, \textbf{I}) \in \mathbb{R}^{c \times h \times w}$, through sequential calls to $\mathcal{Z}_\theta$. 
The quality of the generated image improves with the number of steps to finally yield the latent representation $\mathbf{z}_0^n \in \mathbb{R}^{c \times h \times w}$, where we introduce the superscript $n$ to indicates generation at native resolution $h \times w$ (i.e. same as training resolution). 

\noindent \textbf{Inference-time diffuse-denoise:} Our goal is to use the pretrained parametric denoiser $\mathcal{Z}_\theta$, without further finetuning, to generate $\mathbf{z}^m_0 \in \mathbb{R}^{c \times sh \times sw}$ at a higher resolution $m$, $m = sh \times sw $, where $s$ is the target resolution scaling factor.
The naive approach is to directly start from random noise at the target resolution, $\mathbf{z}^m_T \sim \mathcal{N}(0, I) \in \mathbb{R}^{c \times sh \times sw}$. However, this has been repeatedly shown to lead to suboptimal results, with frequent artifacts and object duplication \cite{demofusion,scalecrafter,hidiffusion}. This is illustrated in Fig.~\ref{fig:ablation_all_di}. 

Instead, prior works proposed a test time diffuse-denoise process \cite{song2020denoising, demofusion, hertz2022prompt}. The idea is to start from the output of the denoising process at native resolution, $\mathbf{z}^n_0$ rather than noise, which is then upsampled to the target resolution $m$ to obtain $\tilde{\mathbf{z}}^m_0 = \mathcal{U}(\mathbf{z}^n_0, s)$, where $\mathcal{U}$ denotes an upsampling function.
Next, $T$ forward diffusion steps progressively add noise to the latents $\tilde{\mathbf{z}}^m_{t=1\ldots T}$.
Finally, the backward process denoises from $\tilde{\mathbf{z}}^m_T$ to yield the final output $\mathbf{z}^m_0$. Note that we use $\tilde{\mathbf{z}}$ and $\mathbf{z}$ to refer to the latents generated during diffusion and denoising respectively.

While a standard denoising process as in Eq.~\ref{eq:denoise} could be used, it often leads to inconsistent global structures, as shown in Fig.~\ref{fig:ablation_all_di_star}. 
Instead, the denoising process from Eq.~\ref{eq:denoise} is now defined as:
\begin{equation}
\label{eq:guidance}
p_{\theta} \left( \mathbf{z}^m_{t-1} | f_t(\tilde{\mathbf{z}}^m_t, \mathbf{z}^m_t)\right)
\end{equation} 

\noindent where $f_t(.)$ is tasked with steering the denoising process and improving the consistency between the high-res and low-res images. Previous work \cite{demofusion,accdiffusion} define $f_t(.)$ as a simple weighted linear combination of $\tilde{\mathbf{z}}^m_t$ and $\mathbf{z}^m_t$ and coin the mechanism \textit{skip residual}. We show in Fig.~\ref{fig:ablation_all_sr} that this yields to suboptimal results. In contrast, we propose a Frequency Modulated approach to defining $f_t(.)$.

\subsection{Frequency-Modulated Denoising} 
\label{ssec:feature-aiding}

The conditioning of the denoising steps through the skip residual has been shown to improve consistency between low and high-resolution images. We however observe that it lacks control over the information transferred. More specifically, the goal of the test-time diffuse-denoise process is to take the upsampled low-resolution image and to produce an output that 1) preserves the global structure, and 2) improves the texture and high-frequency details. The skip residual mechanism however steers the output towards the input indiscriminately, which serves the first objective but can negatively impact the latter.
It would be desirable to instead harness the global structure information from the diffused latents of the forward process, while allowing the denoising process to handle the generation of details. To this end, we appeal to the frequency domain, where global structure and finer details are captured by low- and high-frequency, respectively \cite{marr1980theory,wandell1995foundations, xu2020learning}, and re-define the function $f_t(.)$, which controls information transfer from the forward diffusion into the denoising process, in accordance. 

Let $\mathcal{K}(t)$ be a high-pass filter for timestep $t$, the function $f_t(.)$ in Eq.~\ref{eq:guidance} is defined as follows:
\begin{equation}
\begin{aligned}
    f_t(\tilde{\mathbf{z}}_t^m,\mathbf{z}_t^m) = & IDFT_{2D} ( \mathcal{K}(t) \odot DFT_{2D}\left( \mathbf{z}_t^m \right) \\
    & + (1-\mathcal{K}(t)) \odot DFT_{2D}\left( \tilde{\mathbf{z}}_t^m \right) ),
\end{aligned}
\label{eq:f_mix}
\end{equation}
where $\odot$ denotes the Hadamard product. Essentially, the high-frequency coefficients of the denoised latent $\mathbf{z}_t^m$ are combined with the low-frequency coefficients of the diffused latent $\tilde{\mathbf{z}}_t^m$, modulated by the filter $\mathcal{K}(t)$. Eq.~\ref{eq:f_mix} can be further reformulated in the time domain as below:
\begin{equation}
\label{eq:f_mix_reform}
    f_t(\tilde{\mathbf{z}}_t^m,\mathbf{z}_t^m) = \mathbf{z}^m_t + \mathcal{\kappa}(t) \circledast \bigl( {\tilde{\mathbf{z}}}_t^m - {{\mathbf{z}}}^m_t \bigr),
\end{equation}
where $\mathcal{\kappa}(t) = IDFT_{2D} \left(1 - \mathcal{K}(t) \right) \in \mathbb{R}^{sh \times sw}$ is a convolutional kernel, and $\circledast$ denotes the circular convolution operator. Eq.~\ref{eq:f_mix_reform} shows that the frequency modulation adds a low-frequency update to the denoised latent ${\mathbf{z}}^m_t$ directed towards the diffused latent ${\tilde{\mathbf{z}}}_t^m$, subsequently preserving the global structural information from the upsampled latent. Furthermore, the circular convolution $\mathcal{\kappa}(t)$ in Eq.~\ref{eq:f_mix_reform} can be interpreted as an additional (non-learnable) convolutional layer of the UNet, effectively providing it with a global receptive field and helping generate consistent structure without modifying the UNet architecture~\cite{hidiffusion, fouriscale} or using dilated sampling~\cite{demofusion}.
The result of our FM approach is shown in Fig.~\ref{fig:ablation_all_fm}. In comparison, the skip residual approach of DemoFusion, shown in Fig.~\ref{fig:ablation_all_sr}, produces inconsistencies like a missing left nostril and unnaturally small eyes.

\subsection{Attention Modulation} 
\label{ssec:attention_mixing}
While the FM module successfully maintains global structure and solves the issue of object duplication as shown in Fig.~\ref{fig:ablation_all_fm}, we note that local structures can be inconsistently generated due to the discrepancy between training-time native resolution and the target inference-time high resolutions. 
For example, the top image in Fig.~\ref{fig:ablation_all_fm} shows a distorted mouth compared to the one at native resolution. Similarly, in the bottom example, fur texture is incorrectly generated on the shirt collar. That is, the high-frequency detail generated on the shirt collar is semantically related to one generated on the fox's face and not to the other parts of the shirt. We hypothesize this stems from incorrect attention maps during the high-res denoising stage. This motivates us to propose our Attention Modulation (AM) approach. We take inspiration from attention swapping, a recent method to combine information from two diffusion processes in a more localized manner~\cite{styleprompting,swapanything, photoswap}, and extend the idea to transfer local structural information from the denoising process at native resolution to the one at target resolution.

In particular, the attention of an input tensor $\mathbf{z}$ is computed by first projecting it linearly into a triplet of query, keys, and values, $(Q,K,V)$, respectively, and the self-attention is computed as:

\begin{equation}
    \label{eq:attention}
    Att(\textbf{z}) = softmax \left( \frac{Q \cdot K^T}{\sqrt{d}} \right) V = M \cdot V
\end{equation}

\noindent where $d$ indicates the feature dimensionality, and we refer to $M$ as the attention matrix.

In our case, we modify the self-attention at specific layers of the UNet of the high-resolution denoising process to incorporate information from the attention maps of the native resolution as:

\begin{equation}
    \label{eq:att_mixing}
    \bar{M}^m = (\lambda \cdot \mathcal{U}(M^n, s) + (1-\lambda) \cdot M^m)
\end{equation}

\noindent where $M^n$ and $M^m$ are the attention matrices at native and target resolution respectively, $\lambda$ is a hyperparameter, and $\mathcal{U}$ is an $s$-times upsampling function. The new attention matrix $\bar{M}^m$ is then used instead of $M^m$ during the high-res denoising process in Eq.~\ref{eq:attention}.

Applying our AM module at all layers of the UNet can lead to suboptimal performance due to over-regularization. We apply it instead only for layers in up-blocks of the UNet, as they are known to preserve layout information better~\cite{styleprompting}. Furthermore, we experimented with AM at various stages and found the highest benefit to be at \emph{up\_block\_0}. Results shown in Fig.~\ref{fig:ablation_all_fm_am} demonstrate the benefit of the proposed AM module, particularly regarding better preservation of local structures such as the mouth and shirt collar, highlighted in yellow boxes.
\section{Experiment}
\subsection{Experimental setup}
\label{ssec:experimental_setup}
To demonstrate the effectiveness of our approach, we pair it with a well-performing diffusion model like SDXL \cite{sdxl}. 
For completeness, we also pair our approach with the recent HiDiffusion \cite{hidiffusion}, 
which specifically changes the attention mechanism of SDXL with windowed attention to improve the model latency. 
SDXL is trained at 1024×1024 resolution, which we refer to as \(1\times\).  
We experiment with three unseen higher resolutions such that the model generates $2\times2$, $3\times3$, and $4\times4$ times more pixels than the training setup. 
In the supplementary, we also include results with various aspect ratios, e.g. $2\times 4$, and also experiment with different variants of Stable Diffusion (SD); namely, SD 1.5 \cite{ldm_cvpr22}, SD 2.1 \cite{ldm_cvpr22}, which generate at 512×512 and 768×768 pixels respectively.

\vspace{-2mm}

\noindent \paragraph{Evaluation set.}
Following previous work \cite{demofusion,scalecrafter,accdiffusion,fouriscale} we evaluate performance on a subset of the Laion-5B dataset \cite{laion-5b}. Given the number of compared methods and significant computational demands associated with the task, we randomly sample 10K images from Laion-5b which we use as our real images set, and we sample 1K captions, which we use as text prompts for the models. 

\vspace{-2mm}

\noindent \paragraph{Evaluation metrics.}
Following prior work, we evaluate the quality and diversity of the generated images using Frechet Inception Distance (FID)~\cite{heusel2017gans} and Kernel Inception Distance (KID)~\cite{binkowski2018demystifying}, computed between the generated and real images. Since FID requires resizing images to \(299 \times 299\), which negatively impacts the assessment, it is typical to adopt their patch-level variants~\cite{demofusion, hidiffusion, accdiffusion,fouriscale}. Specifically, we extract 10 random crops from each image before calculating FID and KID, referring to these metrics as FID$_{\text{c}}$ and KID$_{\text{c}}$.
To further evaluate the semantic similarity between image features and text prompts, we report the CLIP score~\cite{radford2021learning}. 
To measure the efficiency of each method, we compute latencies on a single A40 GPU.
\begin{table*}
    \centering
    \begin{tabular}{l|c|cccccc}
    \toprule
         Method & Scaling Factor  & FID$\downarrow$ & KID$\downarrow$  & FID$_{c}\downarrow$ & KID$_{c} \downarrow$ & CLIP $\uparrow$ & Latency(mins) \\ \hline
         \cline{3-8}
         DemoFusion \cite{demofusion} & \multirow{7}{*}{$2\times2$} &63.24 & 0.0084 & 36.75 & 0.0096  & 32.0  & 2.5  \\
         AccDiffusion \cite{accdiffusion}&  & 59.42& 0.0068 &  37.23 &0.0105 & 31.69 &  2.6 \\
         FouriScale*~\cite{huang2024fouriscale}&  & 78.54& 0.0136 &  40.80 &0.0130  & 29.8 & 2.3  \\
         HiDiffusion \cite{hidiffusion}&  & 78.02& 0.0136 & 51.41 & 0.0139  & 30.5  &\textbf{ 0.6}  \\
         HiDiffusion \cite{hidiffusion} + \modelname &  & 69.61&0.0140 & {34.26} & {0.0084} &{32.32}  & 0.8  \\
         SDXL~\cite{sdxl} &  &59.47 &\textbf{0.0067} &50.54 & 0.0136 & 30.6 & 0.8 \\
         SDXL~\cite{sdxl} + \modelname  &  &\textbf{58.91} &0.0072 & {\textbf{33.96}}  & {\textbf{0.0080}}  & { \textbf{32.35}}& 1 \\
         \hline
         DemoFusion \cite{demofusion} & \multirow{7}{*}{$3\times3$}  &\textbf{ 68.82} & 0.0159 & 40.24 &0.0122  & 32.0 & 8.6 \\
         AccDiffusion \cite{accdiffusion}& &73.47  &0.0210 & 43.64 & 0.014 &31.50  &  10 \\
         FouriScale* \cite{huang2024fouriscale}&  & 73.57& 0.0309 &  65.01 &0.0357  & 28.54 & 6.2  \\
         HiDiffusion \cite{hidiffusion}  & & 112.51 & 0.0325& {68.84} & {0.021} & {28.43}   & \textbf{1.5}\\
         HiDiffusion \cite{hidiffusion} + \modelname & & 76.28 &0.0007 & 36.70 & \textbf{0.010} &  \textbf{32.26} & 1.8  \\
         SDXL~\cite{sdxl} & & 78.41 &0.0136 & 69.40 & 0.0210 & 28.44 & 2.2  \\
         SDXL~\cite{sdxl} + \modelname & & 69.25 & \textbf{0.0007}& {\textbf{36.40}} & {\textbf{0.010}} & {{32.25}} &  2.5 \\
         \hline
         DemoFusion \cite{demofusion} & \multirow{7}{*}{$4\times4$} &  65.89 & 0.0087 & 48.44 & 0.0157 & 30.45 &19.6 \\
      AccDiffusion \cite{accdiffusion}&  &73.97 & 0.0090 & 54.80 &  0.0187& 30.15 & 20.5 \\
      FouriScale* \cite{huang2024fouriscale}&  & 105.24& 0.0342 &  70.45 &0.0223  & 27.86 & 14.7  \\
         HiDiffusion \cite{hidiffusion} &  & 129.91 & 0.0483 &  156.98& 0.0877 & 24.32 & \textbf{2.8}\\
         HiDiffusion \cite{hidiffusion} + \modelname &  & 59.05 & 0.0074 & 44.65 &0.0134  & 32.31 &3.1 \\
         SDXL~\cite{sdxl} &  & 160.10 & 0.0602 & 74.37 & 0.0242 &  26.70& 5.4\\
         SDXL~\cite{sdxl} + \modelname &  & \textbf{58.91} & \textbf{0.0073} & \textbf{43.65} & \textbf{0.0130} & \textbf{32.33} & 6.1 \\

    \bottomrule
    \end{tabular}
    \caption{System-level comparisons with SDXL. * indicates inference with FreeU~\cite{si2024freeu}}
    \label{tab:comp_to_sota}
\end{table*}

\subsection{Main Results}

\begin{figure}[t] 
    \centering
    \subfloat[Low Resolution Att.]{
        \includegraphics[width=0.31\linewidth]{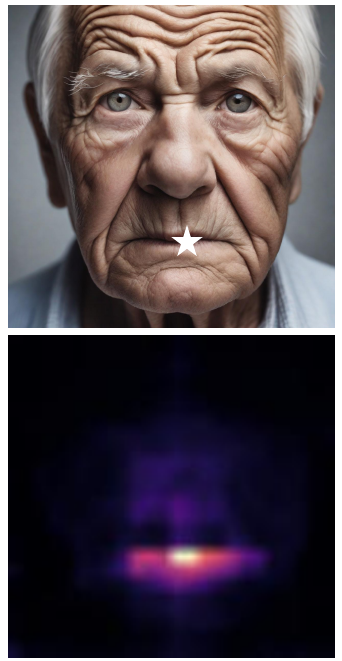}
    }
    \subfloat[High Resolution Att.]{
        \includegraphics[width=0.31\linewidth]{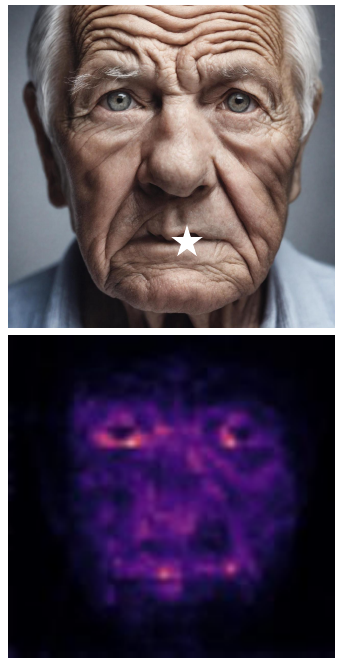}
    }
    \subfloat[Att. Modulation]{
            \includegraphics[width=0.31\linewidth]{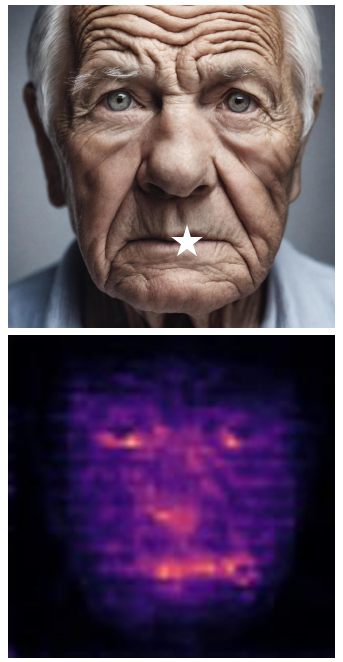}
    }
    \caption{Visualization of Attention Maps in the UNet: (a) Low-Resolution Attention map, (b) High-Resolution Attention map, (c) Attention Map when using the AM module}
    \label{fig:attention_map}
\end{figure}

\begin{figure}[t] 
    \centering
    \subfloat[Direct Upsampling]{
        \includegraphics[width=0.31\linewidth]{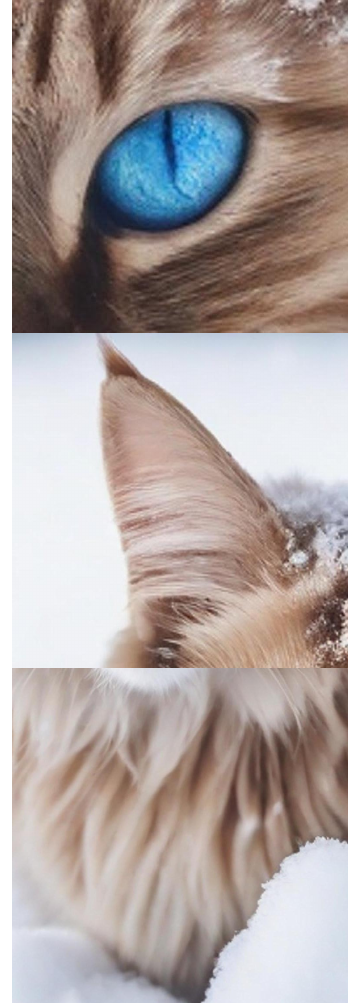}
    }
    \subfloat[BSRGAN]{
        \includegraphics[width=0.31\linewidth]{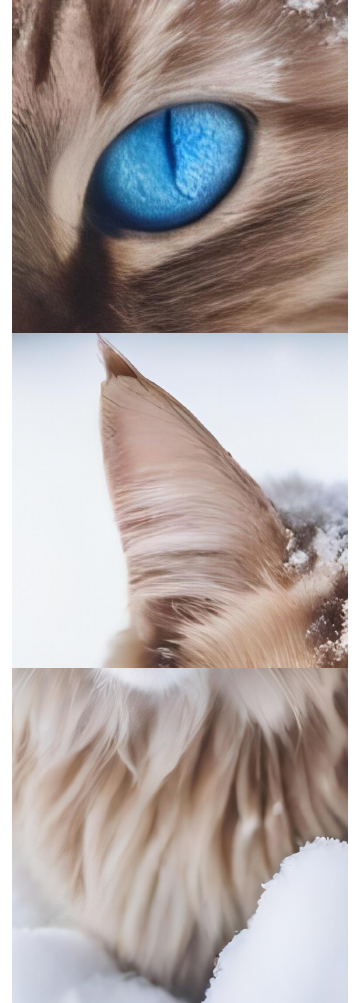}
    }
    \subfloat[Ours]{
        \includegraphics[width=0.31\linewidth]{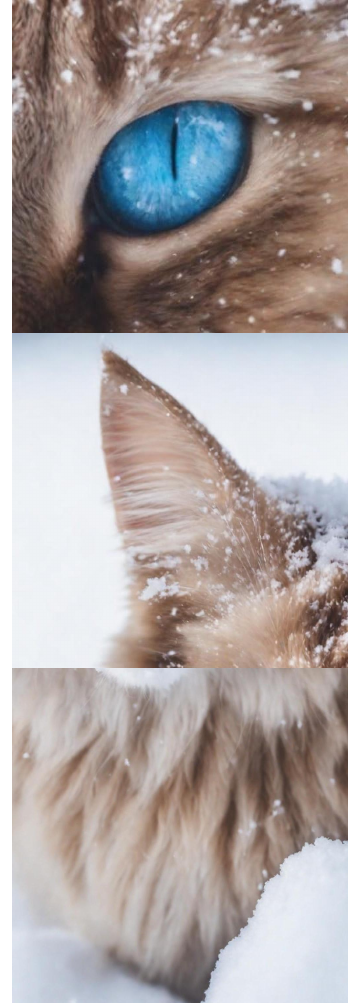}
    }
    \caption{Qualitative comparison between Direct Upsampling, BSRGAN, and our method. The patches shown were cropped from a $4096 \times 4096$ resolution image. 
    Zoom in for best view.}
    \label{fig:sr_upsample_comparison}
\end{figure}

\begin{figure*}[]
\centering
\subfloat[Native Resolution Image]{
    \includegraphics[width=0.95\textwidth]{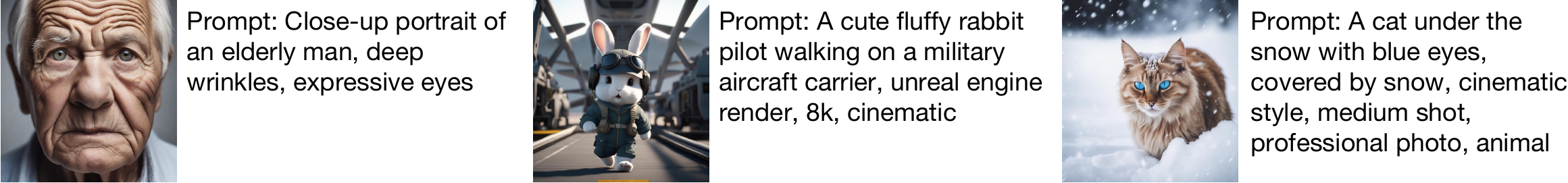}
} \\
\subfloat[DemoFusion]{
    \includegraphics[width=\textwidth]{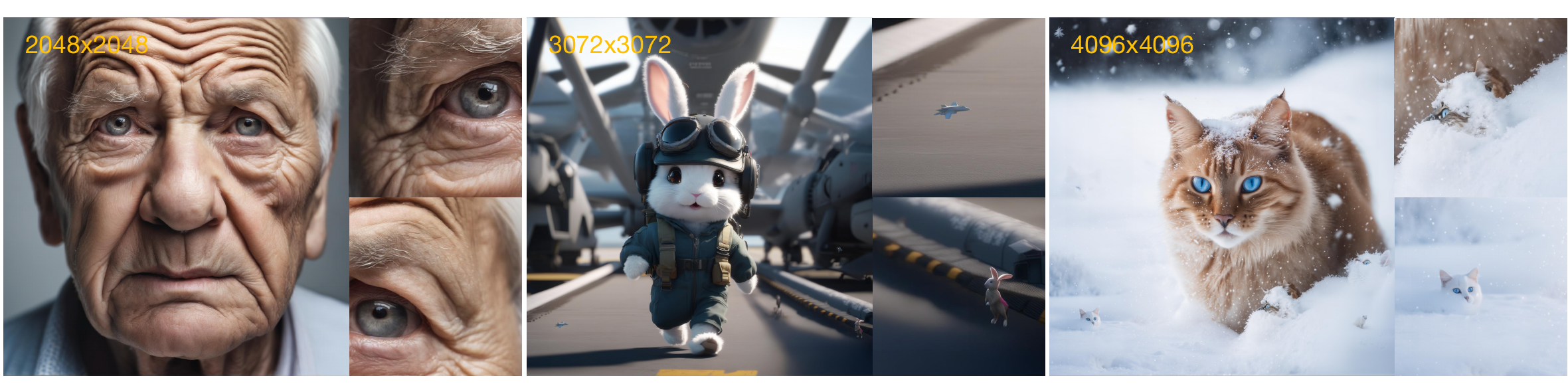}
} \\
\subfloat[FouriScale*]{
    \includegraphics[width=\textwidth]{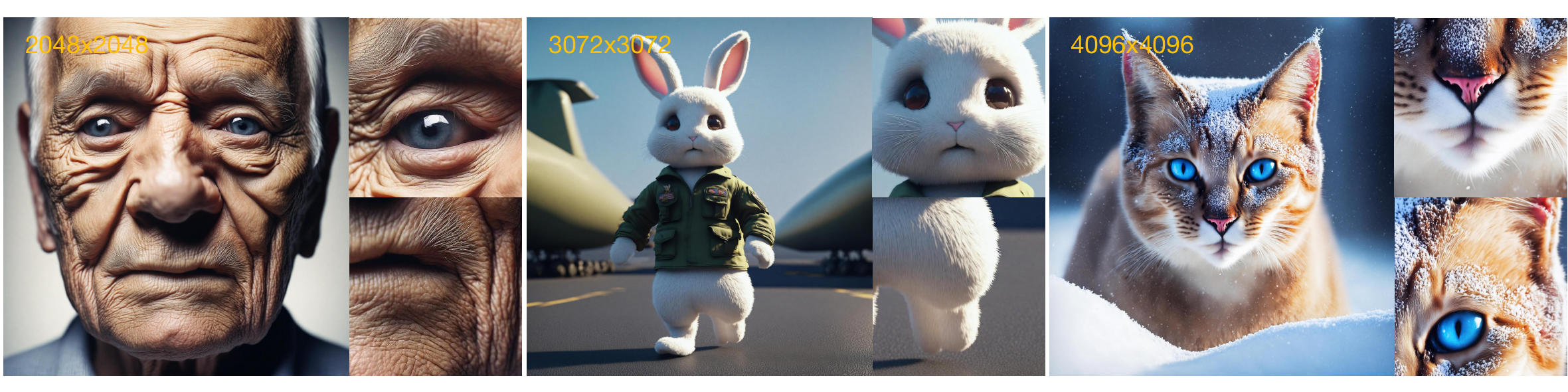}
} \\
\subfloat[HiDiffusion]{
    \includegraphics[width=\textwidth]{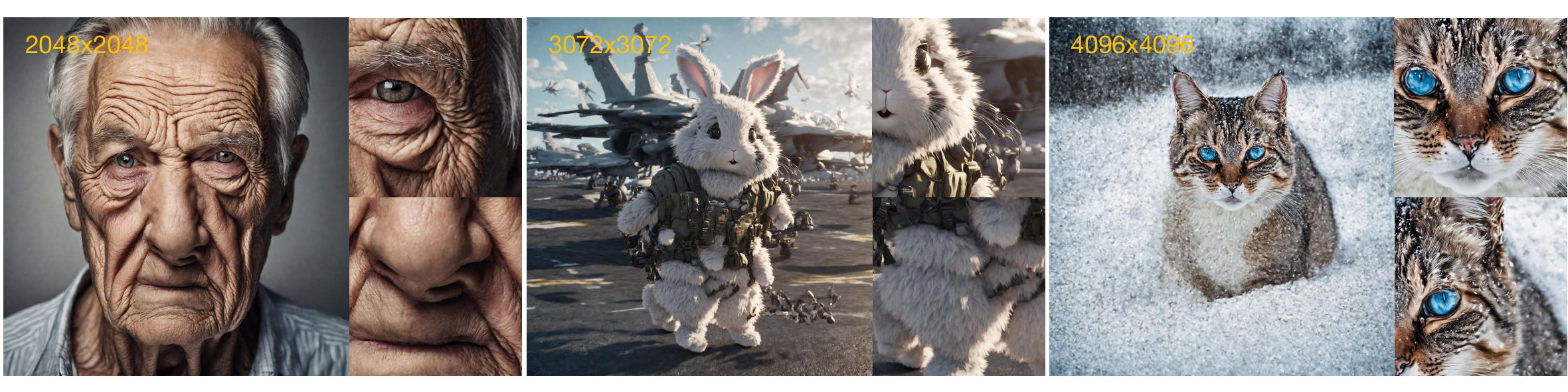}
} \\
\subfloat[Our Method]{
    \includegraphics[width=\textwidth]{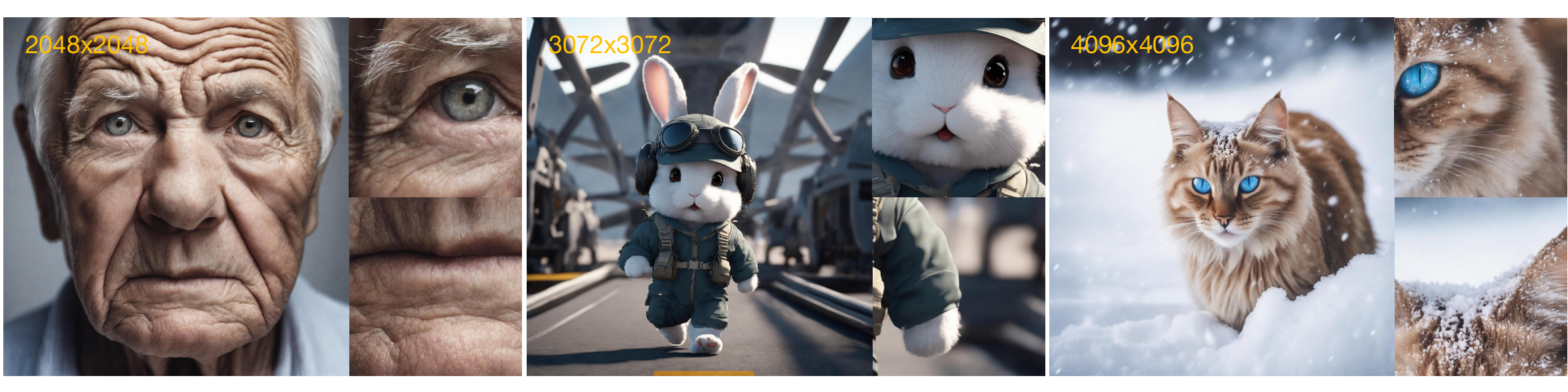}
}
\caption{Qualitative comparison with other methods based on SDXL. Best viewed when zoomed in. * indicates inference with FreeU~\cite{si2024freeu}}
\label{fig:comparison_qualitative}
\end{figure*}

\begin{figure}[t] 
    \centering
    \subfloat[Constant LF]{
        \includegraphics[width=0.46\linewidth]{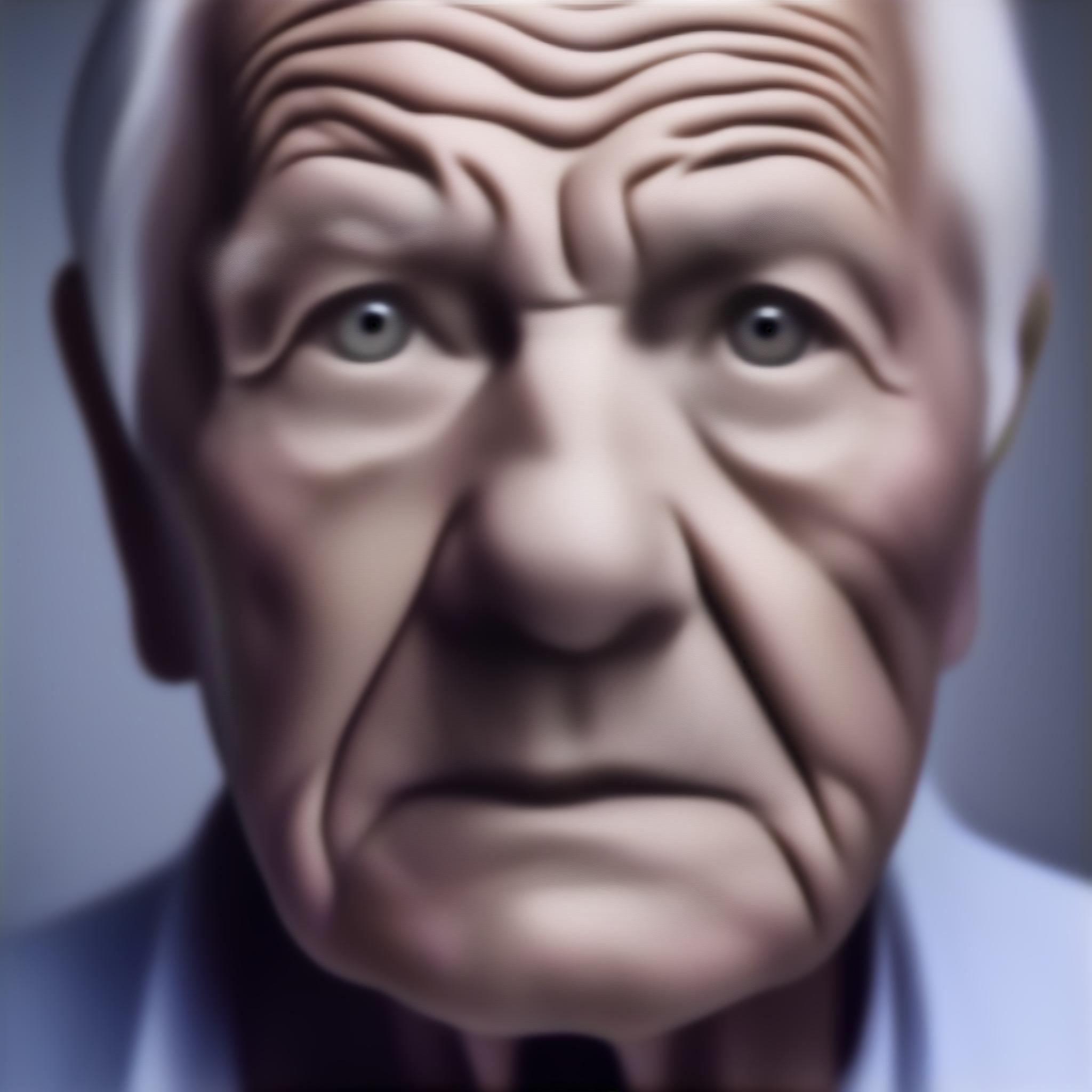}
        \label{fig:constant_lf}
    }
    \subfloat[Time-aware LF]{
        \includegraphics[width=0.46\linewidth]{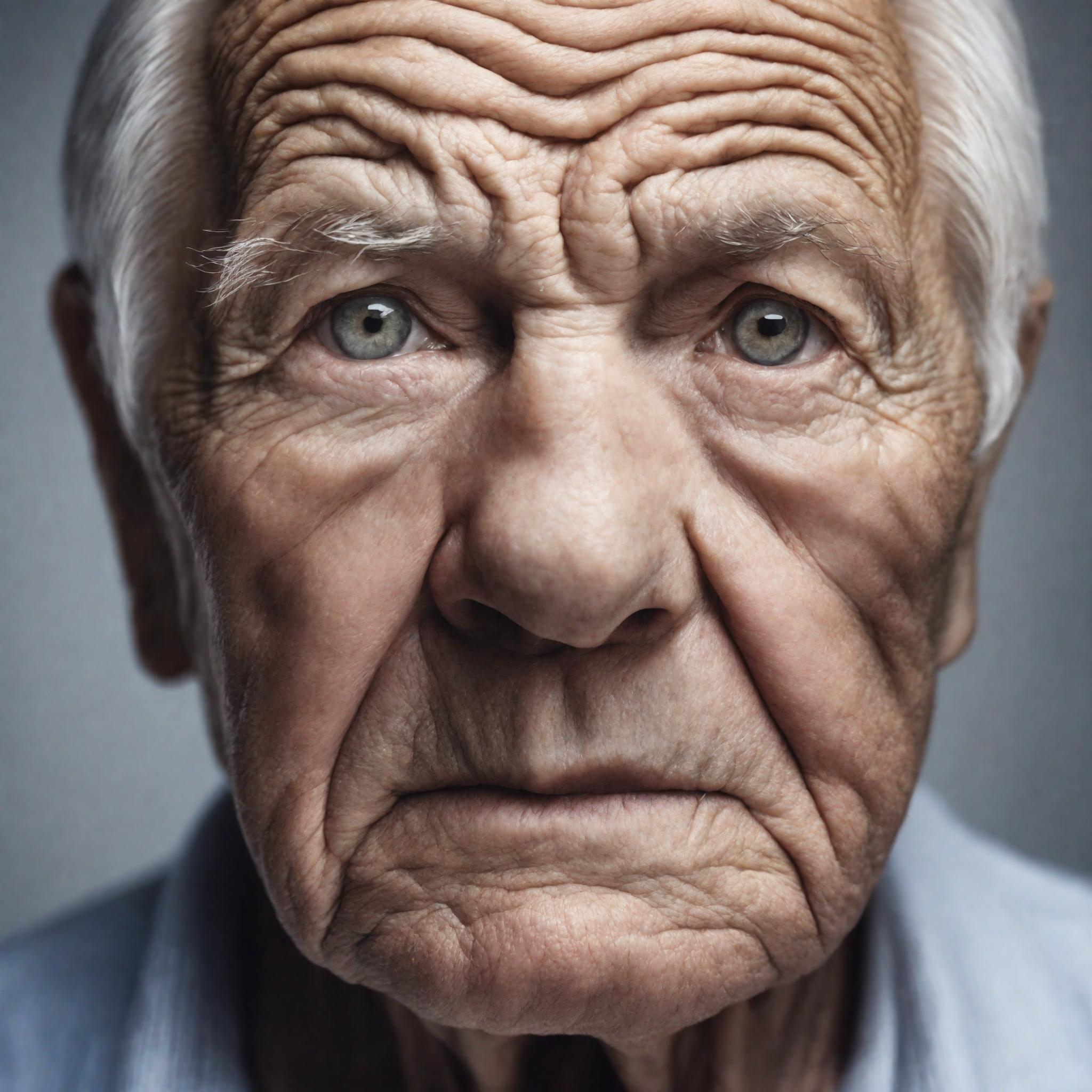}
        \label{fig:time_aware_lf}
    }
    \caption{Comparison between Constant LF and Time-aware LF.}
    \label{fig:lf_comparison}
\end{figure}
We select Demofusion~\cite{demofusion}, AccDiffusion~\cite{accdiffusion}, FouriScale~\cite{fouriscale}, and HiDiffusion~\cite{hidiffusion} as representative methods of the current state-of-the-art among high-resolution generation methods.
As shown in Table~\ref{tab:comp_to_sota}, \modelname{} achieves the best overall performance on $\text{FID}_{c}$, $\text{KID}_{c}$, and CLIP Score in all cases. 
In the case of $\text{FID}$ and $\text{KID}$, \modelname{} provides substantial gains for larger scale factors, while producing similar results to DemoFusion on lower scale factors. However, these metrics heavily downsample high-resolution images before computing the metrics and thus do not capture finer details in the evaluation results. This is a widely-known issue for these metrics, as explained in Sec.~\ref{ssec:experimental_setup}. Finally, we note that our method adds only small latency overheads compared to direct inference on the target resolution, e.g. 0.2, 0.3, and 0.7 min at $2\times$, $3\times$ and $4\times$ scale factors respectively when combined with SDXL. In comparison, DemoFusion adds 14.2 sec latency vs SDXL direct inference at $4 \times$ scale factor.
When compared to the frequency-based method FouriScale~\cite{fouriscale}, \modelname{} also shows notable improvements in both quality and latency. For instance, under 4K resolution image generation, it achieves \textbf{43.65} vs. \textbf{70.45} on $\text{FID}_{c}$ and \textbf{32.31} vs. \textbf{26.67} on CLIP score, while also being faster than FouriScale.
Additionally, we observed that \modelname{} can be seamlessly integrated into single-pass methods, such as HiDiffusion~\cite{hidiffusion}, to enhance performance while maintaining fast image generation, achieving an effective latency-quality trade-off.
These results quantitatively validate the effectiveness of our method in improving the quality of image generation. 

In Figure~\ref{fig:comparison_qualitative}, we present a comparison between DemoFusion, FouriScale, HiDiffusion, and \modelname{}. We selected three complex textual prompts to highlight the image-generation capabilities of the model. For FouriScale, we used the default setting with FreeU~\cite{si2024freeu}.
Firstly, as mentioned above, DemoFusion tends to generate repetitive content and artifacts with unreasonable local structures due to its patch-based generation approach (see for example the two small cat heads generated on the top-right image). FouriScale~\cite{fouriscale} and HiDiffusion~\cite{hidiffusion} produce visually unappealing structures and extensive areas of irregular textures, which significantly degrade the overall visual quality. 
Additionally, we compare our method with the super-resolution approach BSRGAN~\cite{zhang2021designing}, as shown in Figure~\ref{fig:sr_upsample_comparison}. 
We observe that \modelname{} effectively introduces or modifies high-frequency details that were not present in the original image, while preserving structural information, leading to more appealing and detailed images.

To further illustrate the generality of our approach, in the supplementary material we provide results of our approach in combination with SD1.5 and SD2.1.

\subsection{Ablation Study}
In this section, we conduct ablation studies and use SDXL with the $2\times2$ scale factor setting. 

\vspace{-8mm}

~\paragraph{Effectiveness of the components in the \modelname{}}
We study the effect of the two components of \modelname{}, Frequency-Modulated Denoising (FM) and Attention Modulation (AM). The results shown in Figure~\ref{fig:ablation_all} indicate the following:
(1) both direct inference from random noise, and direct inference from the diffused latent at native resolution generate outputs with structural distortions and repeated patterns.  
(2) while the Skip Residuals of DemoFusion helps maintain the global structure of the image, it still produces artifacts and poor local patterns.  
(3) Compared to Skip Residuals, FM reduces undesirable local patterns by leveraging the low-frequency information of the image at native resolution, which provides better structural guidance. 
(4) Attention Modulation resolves inconsistencies between local patterns and global structure by utilizing the attention map from the native resolution, offering strong guidance of the semantic relationships among latent tokens.  
Overall, FM and AM address structural distortions and local pattern inconsistencies in high-resolution images effectively, highlighting the meaningful contributions of \modelname{}.

\vspace{-8mm}

~\paragraph{Effectiveness of the time-aware formulation on the FM module}
We show here the effect of the time-varying formulation of FM, as illustrated in Figure~\ref{fig:constant_lf}. Specifically, the FM module incorporates low-frequency information from the corresponding diffused latent at each step $t$. Instead, we can avoid this time-varying nature and utilize the upsampled latent as a single static reference. However, this approach results in images that appear noticeably blurrier and lose finer details associated with high-frequency information, highlighting the importance of the dynamic nature of the FM module throughout the denoising process.

\vspace{-8mm}

~\paragraph{Analysis of Attention Modulation}
 To better understand the principles underlying the AM module, we visualize in Figure~\ref{fig:attention_map} the self-attention maps of a tokens from the mouth region (marked with a star) as the query and all tokens as the key and value. 
 The resulting attention map computed using the low-resolution latent primarily encodes coarse information of the semantic relations among parts of the image, but lacks fine-grained contextual information across the entire face. Instead, the attention maps at high resolution are more detailed, but fail to capture semantic relatedness, e.g. the mouth areas are not highlighted.  
 After applying AM, the attention map effectively integrates local-global relationships with enhanced fine-grained detail. This analysis provides visual insights into how AM repairs inconsistencies in local patterns, contributing to more coherent global structures.

\section{Conclusion}
We introduced \modelname{}, a training-free diffusion model for high-resolution image generation. 
To address issues of object repetition and structural distortion, we propose a \textit{Frequency Modulated} strategy. 
By leveraging the Fourier domain, this method enhances guidance for high-resolution generation while avoiding latency overheads associated with multi-patch approaches.
Additionally, we propose an effective \textit{Attention Modulation} mechanism to address inconsistent local texture patterns, a challenge largely overlooked in previous works.
Extensive quantitative and qualitative evaluations highlight the effectiveness of our method. We further show that, contrary to previous works, our method incurs in marginal latency overheads.
{
    \small
    \bibliographystyle{ieeenat_fullname}
    \bibliography{main}
}

\clearpage
\appendix

\section{Appendix}

To complement the main content of the paper, we provide here additional details about the method in Sec.~\ref{sec:add-details} as well as additional quantitative and qualitative results in Sec~\ref{sec:add-results}.

\section{Additional technical details}\label{sec:add-details}

\subsection{Frequency Modulation details}

\paragraph{Time-varying high-pass filter definition.}
In our method, we rely on frequency domain and use a high pass filter to steer the denoising process as described in equation \eqref{eq:f_mix}. In the following, we provide the formal definition of the time-varying high pass filter, $\mathcal{K}(t)$, that we used.

The high-pass filters $\mathcal{K}(t)$ have time-varying cut-off frequencies, defined as follows:
\begin{align}
\rho (t) &= \frac{t}{T} \label{eq:s} \\
{\tau _h}(t) &= h \cdot c \cdot (1-\rho (t)) \label{eq:threshold_h} \\
{\tau _w}(t) &= w \cdot c \cdot (1-\rho (t)) \label{eq:threshold_w}
\end{align}
where ${\tau _h}(t)$ and ${\tau _w}(t)$ are the horizontal and vertical cut-off frequencies at timestep $t$, respectively. Subsequently, the mask $\mathcal{K}(t)$, which is applied on the shifted frequency spectrum centered on $(x_c, y_c)$, is defined as
\begin{align}
\mathcal{K}(t) =
\begin{cases} 
\rho (t), & \text{if } 
\left| x - x_c \right| < \frac{{\tau _w}(t)}{2} \\
& \quad \text{\& } 
\left| y - y_c \right| < \frac{{\tau _h}(t)}{2}, \\
1, & \text{otherwise}
\end{cases} \label{eq:k}
\end{align}

The cut-off frequency grows as the denoising process progresses, while the scaling factor of the low-frequency coefficients decreases. Our frequency modulation is designed such that the guidance from the denoised latent ${\tilde{\mathbf{z}}}_t$ becomes more significant as $t \rightarrow 0$. In our experiments, we set $c = 0.5$.

\paragraph{Derivation of the Frequency Modulation in time-domain.}
\begin{figure}[t] 
    \centering
    \subfloat[Swapping]{
        \includegraphics[width=0.5\linewidth]{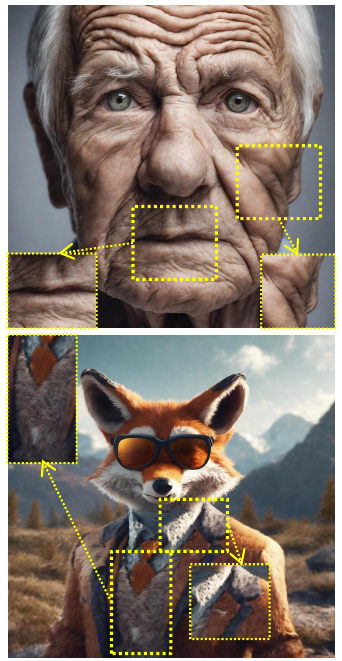}
        \label{fig:swapping}
    }
    \subfloat[Modulation]{
        \includegraphics[width=0.5\linewidth]{figures/figure_all_ablation_FULL_V4.pdf}
        \label{fig:mixing}
    }
    \caption{Comparison of Attention Swapping and Modulation}
    \label{fig:mixing_swapping}
\end{figure}
In the main paper, we mention that our frequency modulation introduced in Eq.~\eqref{eq:f_mix} can be reformulated in time domain as Eq.~\eqref{eq:f_mix_reform} and discuss the corresponding benefits. Here, we provide a formal derivation to support the equivalence between the two formulations. For ease of presentation, we omit the timestep $t$ and resolution $m$ notations from operands. 

Let $\mathbf{z} \in \mathbb{R}^{h \times w}$ be the 2D latent, and $\mathbf{Z} = DFT_{2D}\left( \mathbf{z} \right) \in \mathbb{C}^{h \times w}$ be the Fourier transform of $\mathbf{z}$. Written in matrix form, 
\begin{equation}
    \mathbf{Z} = ({W_r}\mathbf{z}{W_c}),
\end{equation}
where ${W_r} \in {\mathbb{C}^{h \times h}},{W_c} \in {\mathbb{C}^{w \times w}}$ are the row- and column-wise Fourier transform matrices, respectively. Let $\mathcal{K} \in \mathbb{R}^{h \times w}$ be the high-pass filter defined in the previous section, our proposed mixing operation in the frequency domain is formulated as below:
\begin{align*}
    \hat{\mathbf{Z}} & = \mathcal{K} \odot DFT_{2D}(\mathbf{z}) + (1 - \mathcal{K}) \odot DFT_{2D}(\tilde{\mathbf{z}}) \\ 
           & = \mathcal{K} \odot ({W_r}\mathbf{z}{W_c}) + (1 - \mathcal{K}) \odot ({W_r}\tilde{\mathbf{z}}{W_c}) \\
           & = {W_r}\mathbf{z}{W_c} + (1 - \mathcal{K}) \odot \left({W_r}(\tilde{\mathbf{z}} - \mathbf{z}){W_c} \right)
\end{align*}

The inverse DFT of $\hat{\mathbf{Z}}$, which is the outcome of Eq.~\ref{eq:f_mix}, is formulated as:
\begin{align*}
    \hat{\mathbf{z}} & = IDFT_{2D}(\hat{\mathbf{Z}})\\
 & = W_r^{ - 1}\left({W_r}\mathbf{z}{W_c} + (1 - \mathcal{K}) \odot \left({W_r}(\tilde{\mathbf{z}} - \mathbf{z}){W_c} \right) \right)W_c^{ - 1}\\
 & = W_r^{ - 1}{W_r}\mathbf{z}{W_c}W_c^{ - 1} \\ & \quad \quad + W_r^{ - 1}\left((1 - \mathcal{K}) \odot ({W_r}(\tilde{\mathbf{z}} - \mathbf{z}){W_c}) \right)W_c^{ - 1}\\
 & = \mathbf{z} + \left(W_r^{ - 1}(1 - \mathcal{K})W_c^{ - 1} \right) \circledast \left(W_r^{ - 1}{W_r}(\tilde{\mathbf{z}} - \mathbf{z}){W_c}W_c^{ - 1} \right)\\
 & = \mathbf{z} + k \circledast (\tilde{\mathbf{z}} - \mathbf{z}),
\end{align*}
resulting in Eq.~\ref{eq:f_mix_reform} in the main paper, where $k = W_r^{-1}(1 - K)W_c^{-1} = IDFT_{2D}(1-\mathcal{K})$ is a convolutional kernel and $\circledast$ denotes a circular convolution operator.

\subsection{Attention Modulation analysis}

As mentioned in Sec.~\ref{ssec:attention_mixing}, we take inspiration from recent literature using attention swapping to control local texture. However, rather than swapping attention, we mix the two attention paths instead. In Figure~\ref{fig:mixing_swapping} we compare attention swapping versus our proposed attention modulation. These results clearly show the benefit of including the attention from the high resolution path rather than directly swapping with the low res pass to avoid loss of information from the high res denoising path. We empirically set $\lambda$ used in Eq~\eqref{eq:attention} to $0.7$.

\section{Additional experimental results}\label{sec:add-results}
\subsection{FAM diffusion with different SD backbones}
In Table~\ref{tab:comp_to_sota} we show that our method outperforms several baselines when combined with SDXL. In addition to those main results, we further combine our FAM diffusion method with various SD backbones. The quantitative results in Table~\ref{tab:ours_plus_sdvar-supp} demonstrate that our approach can seamless combine with different variants of SD and provides similarly large improvements in quality and image-text alignment across all experimental settings. 

\subsection{FAM diffusion with different aspect ratios}
\begin{figure*}[]
\centering
\subfloat[Native Resolution Image]{
    \includegraphics[width=0.95\textwidth]{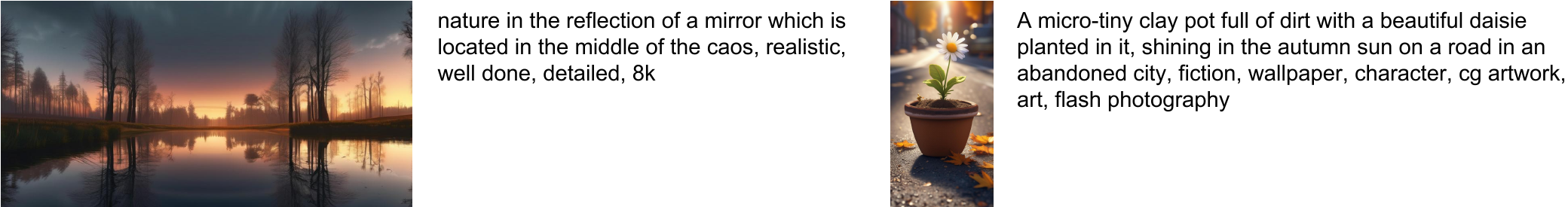}
} \\
\subfloat[DemoFusion]{
    \includegraphics[width=\textwidth]{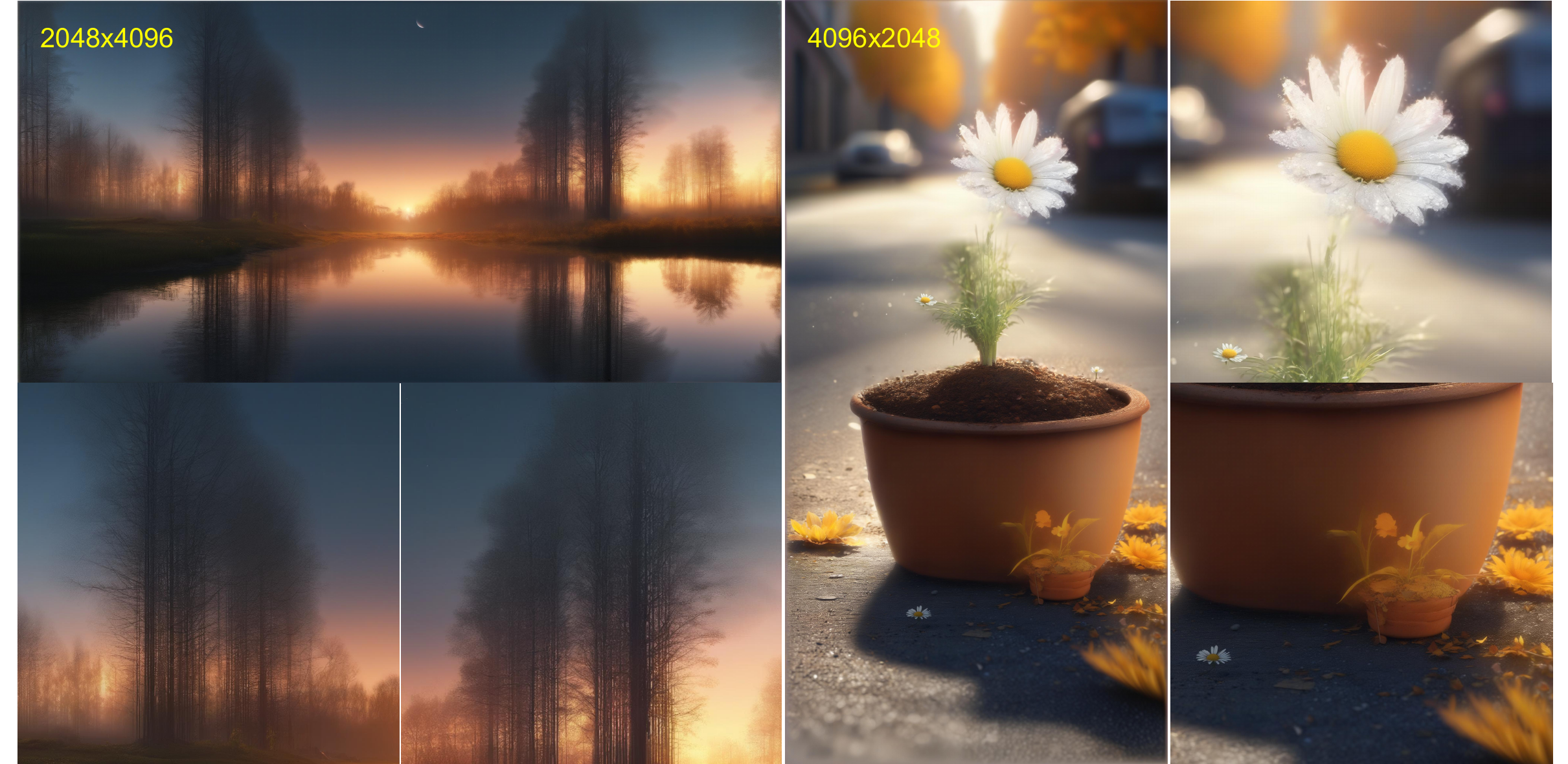}
} \\
\subfloat[FouriScale*]{
    \includegraphics[width=\textwidth]{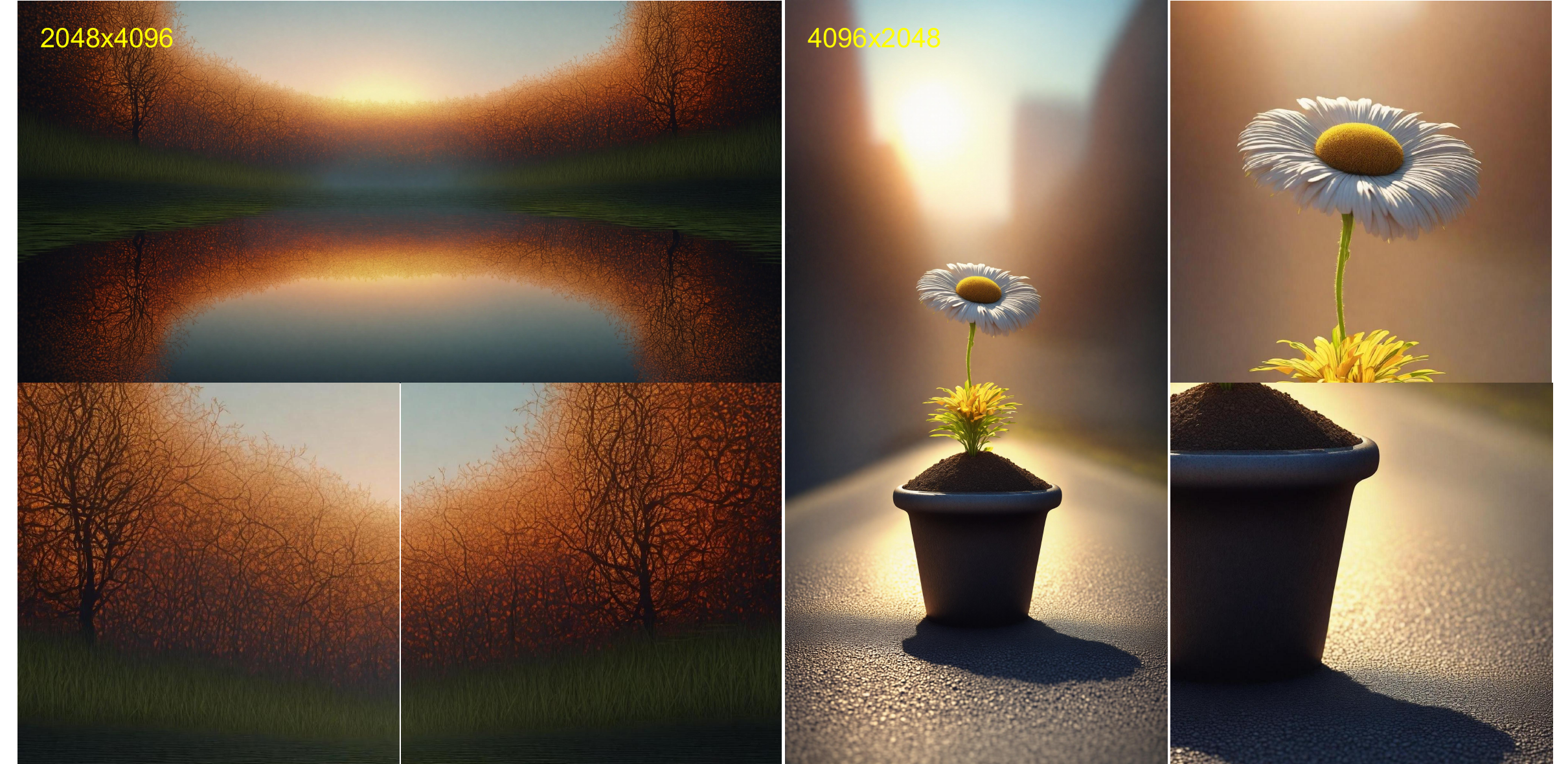}
} \\

\caption{Qualitative comparison with other methods based on SDXL. Best viewed when zoomed in. * indicates inference with FreeU~\cite{si2024freeu}. \textit{(Continued in Fig.~\ref{fig:comparison_qualitative_non_square_2})}.}
\label{fig:comparison_qualitative_non_square_1}
\end{figure*}

\begin{figure*}[]
\centering

\subfloat[HiDiffusion]{
    \includegraphics[width=\textwidth]{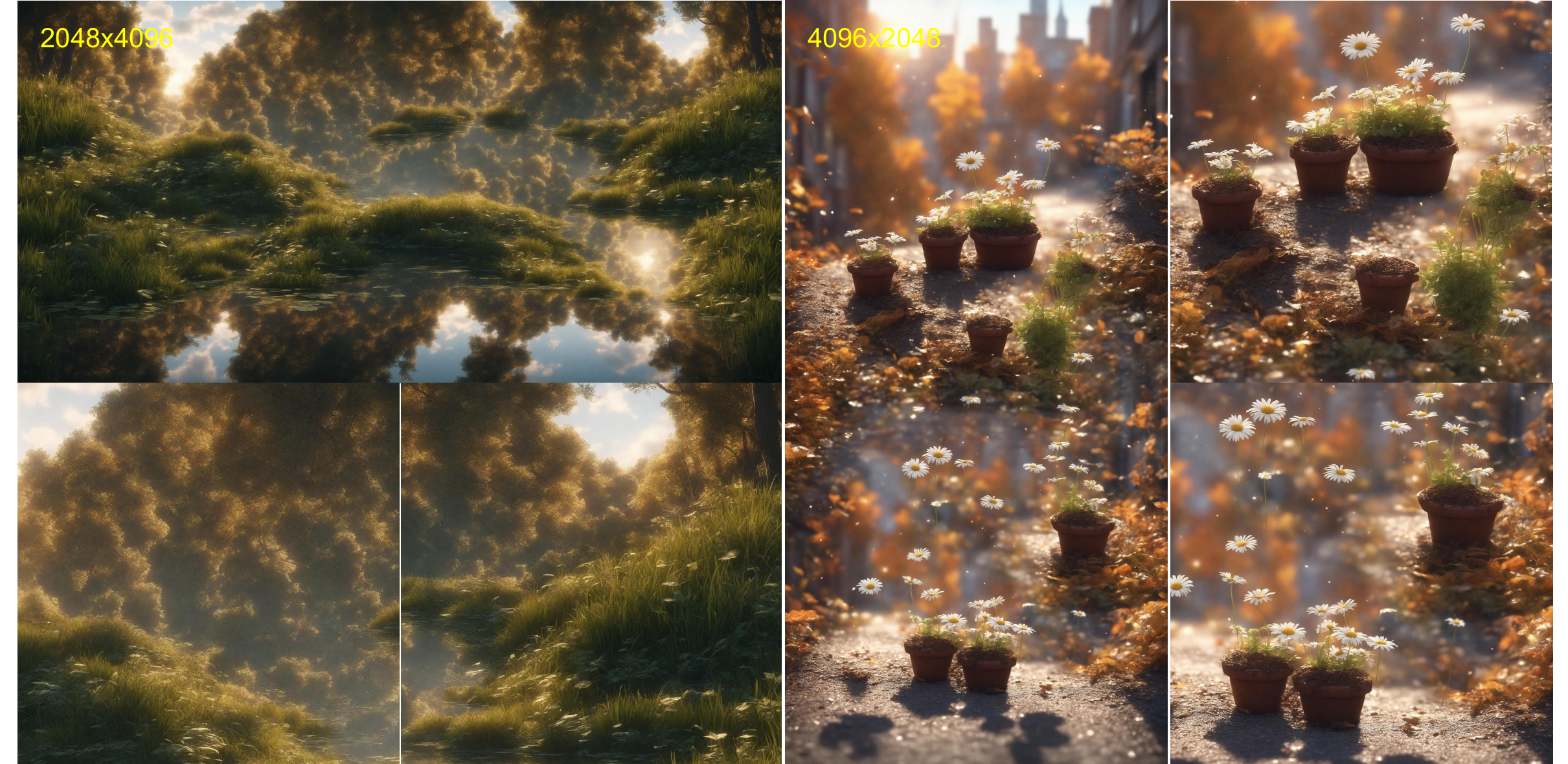}
} \\
\subfloat[Our Method]{
    \includegraphics[width=\textwidth]{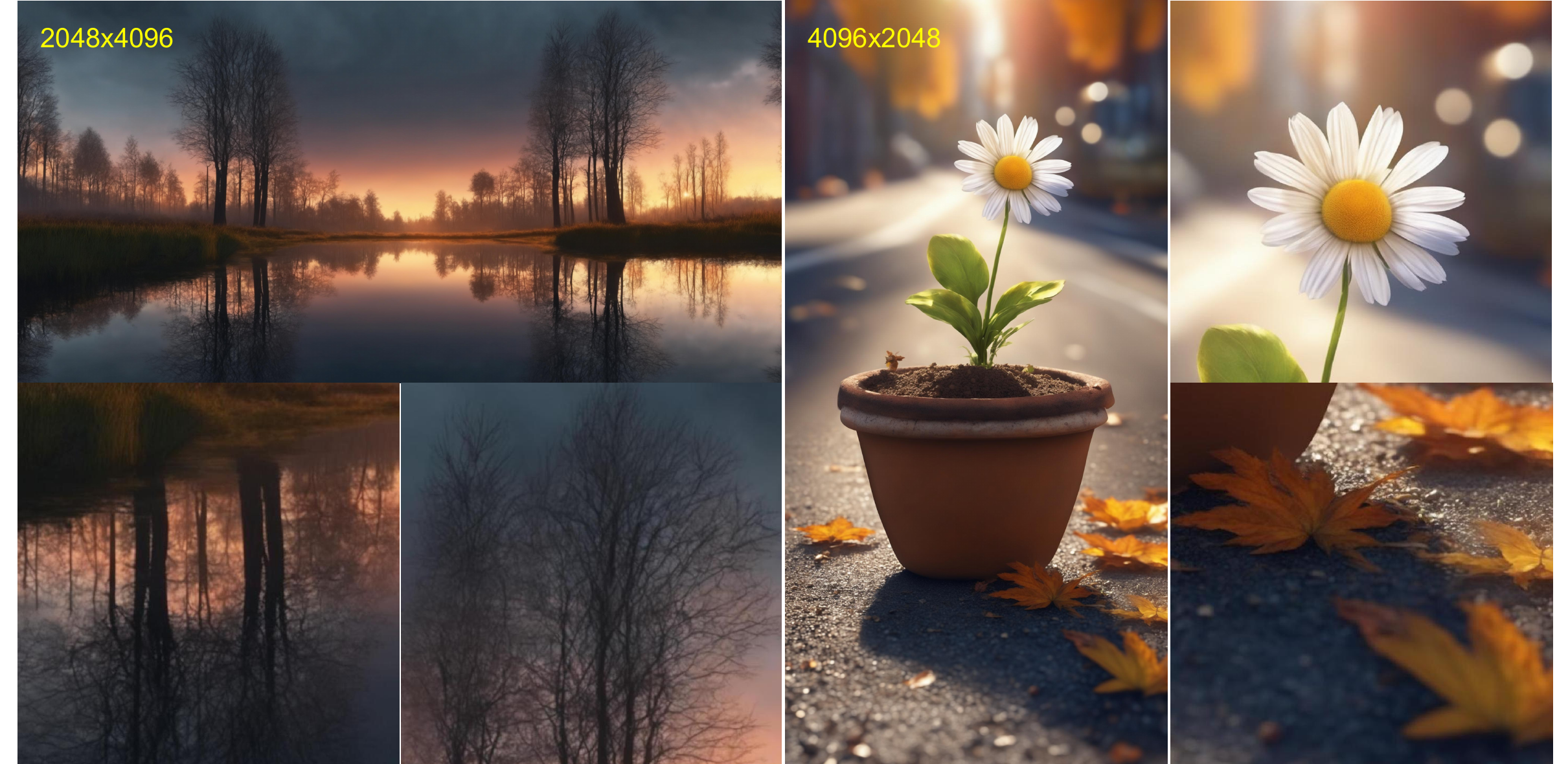}
}
\caption{Qualitative comparison with other methods based on SDXL \textit{(continued from Fig.~\ref{fig:comparison_qualitative_non_square_1})}. Best viewed when zoomed in.}
\label{fig:comparison_qualitative_non_square_2}
\end{figure*}

\begin{figure*}[t] 
    \centering
    
    \subfloat[FouriScale*]{
        \includegraphics[width=0.3\linewidth]{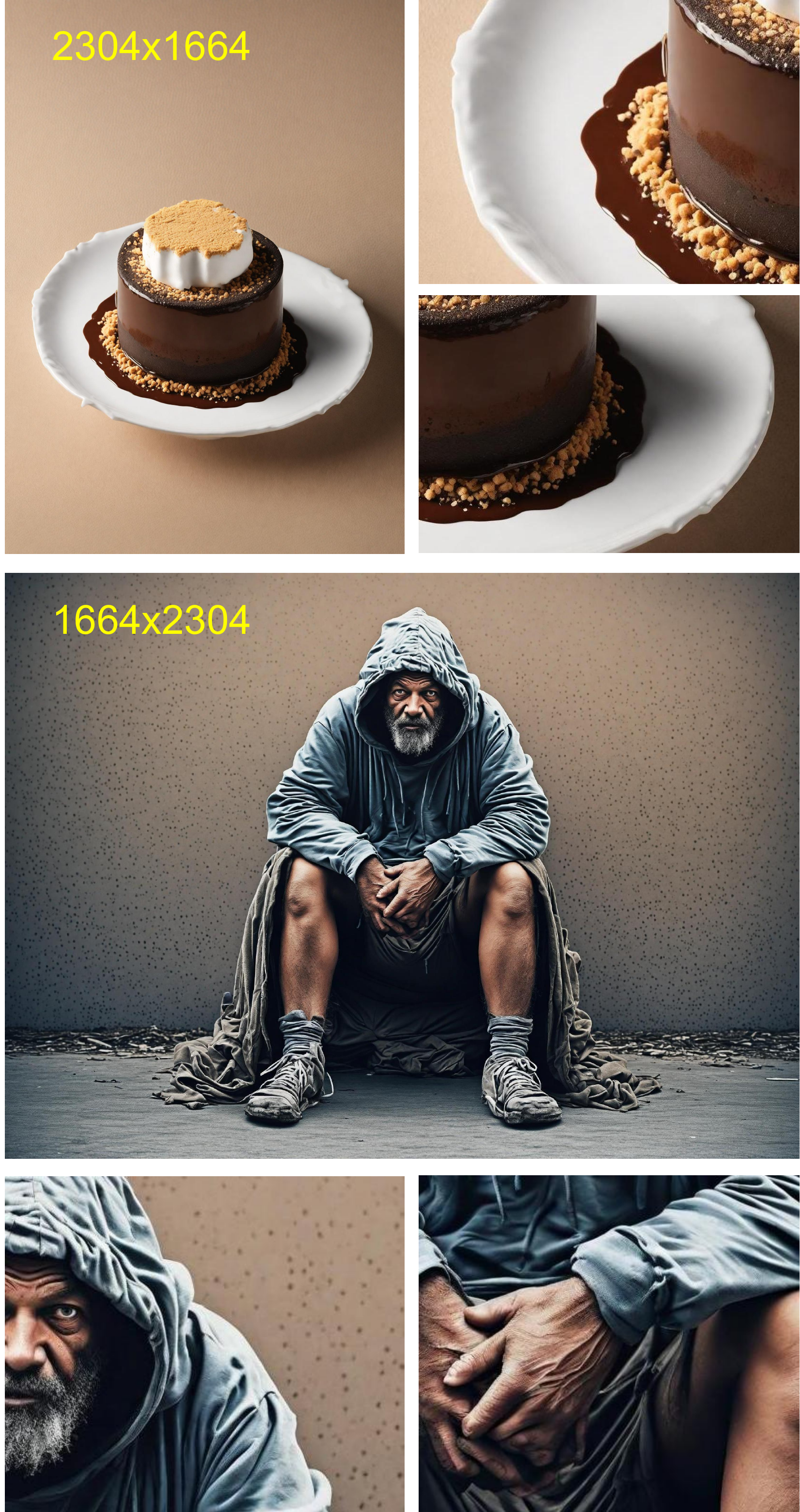}
        \label{fig:non_1024_fouri}
    }
    \subfloat[HiDiffusion]{
        \includegraphics[width=0.3\linewidth]{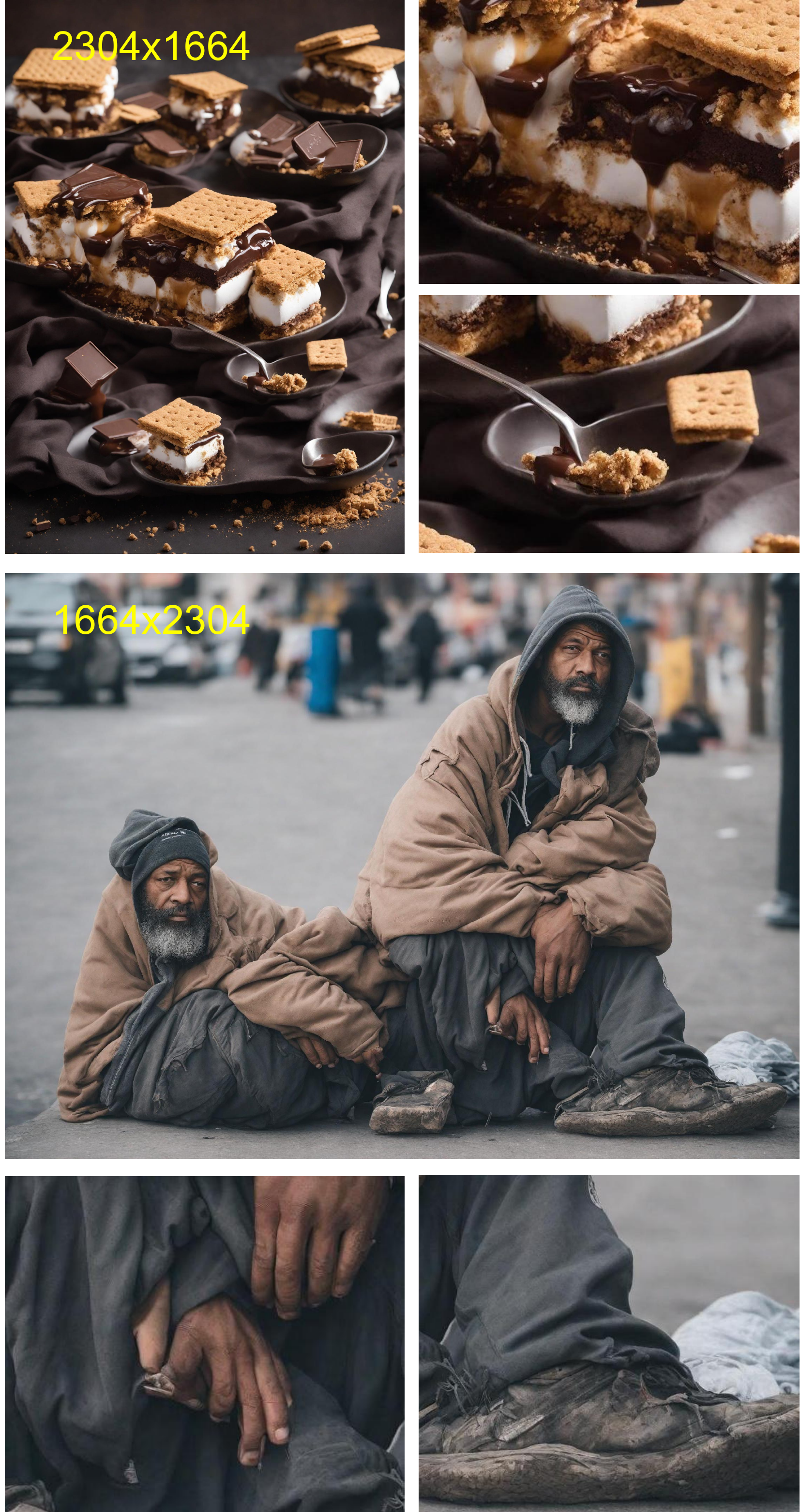}
        \label{fig:non_1024_hidiff}
    }
    \subfloat[Our Method]{
        \includegraphics[width=0.3\linewidth]{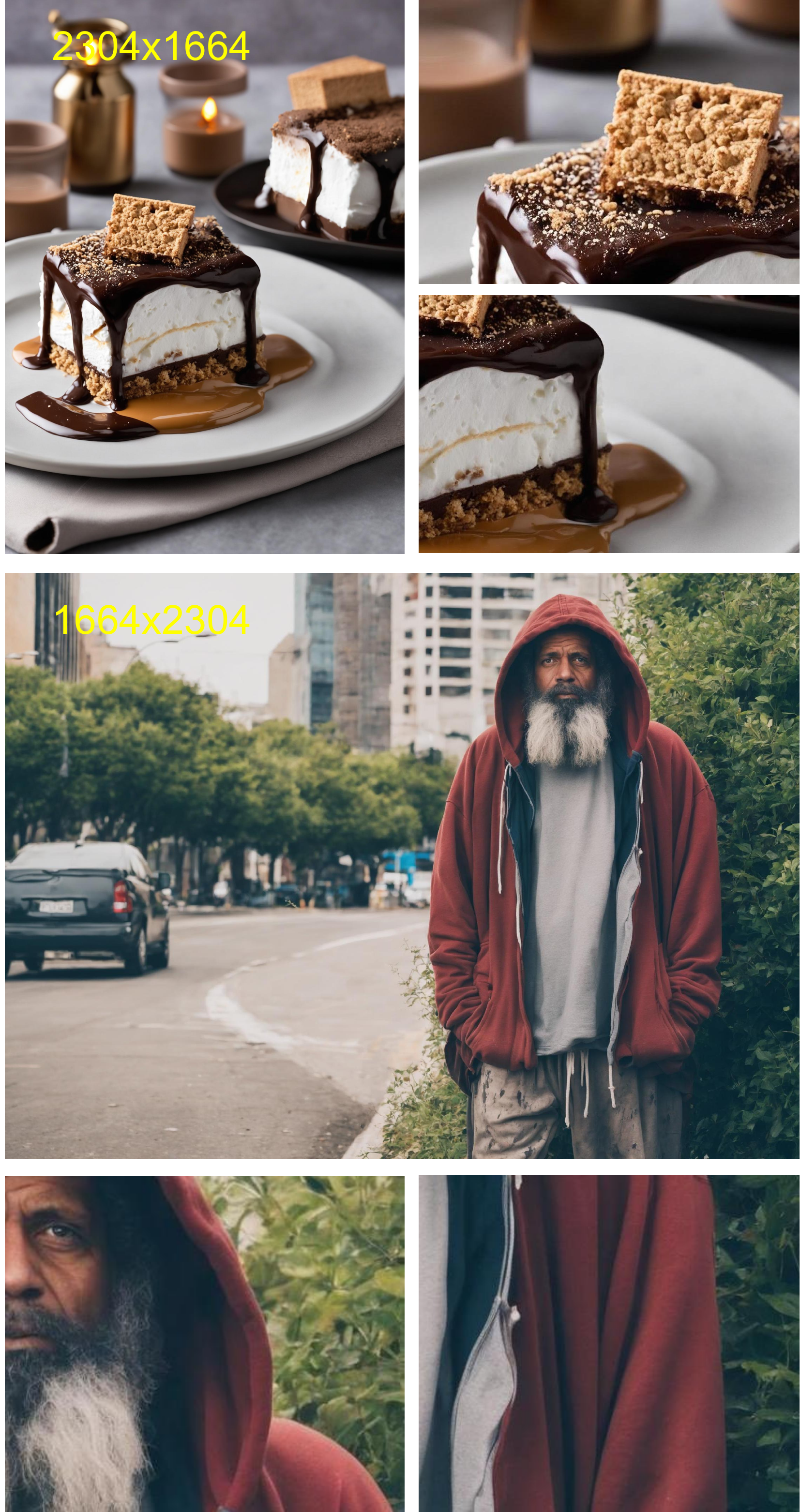}
        \label{fig:non_1024_our}
    }
    \caption{Qualitative comparison with other methods based on SDXL with arbitrary resolutions. DemoFusion is unable to handle arbitrary resolutions, therefore not included. Best viewed when zoomed in.}
    \label{fig:arb_res}
\end{figure*}

Thus far, we have used our method to generate high-resolution images by equally upscaling both the height and width. Here, we study the effect of using Fam diffusion targeting different aspect ratios. In particular, starting from the SDXL model, we use our approach targeting higher resolutions with different aspect ratios. The quantitative results in Table~\ref{tab:comp_to_sota-24} and qualitative results shown in Figures~\ref{fig:comparison_qualitative_non_square_1} through~\ref{fig:arb_res}, clearly highlight the versatility of our method that can seamlessly adapt to various settings without compromising quality. 

\subsection{FAM diffusion with different conditioning terms}
\begin{figure*}
\centering

\subfloat[Image to Image]{
    \includegraphics[width=\textwidth]{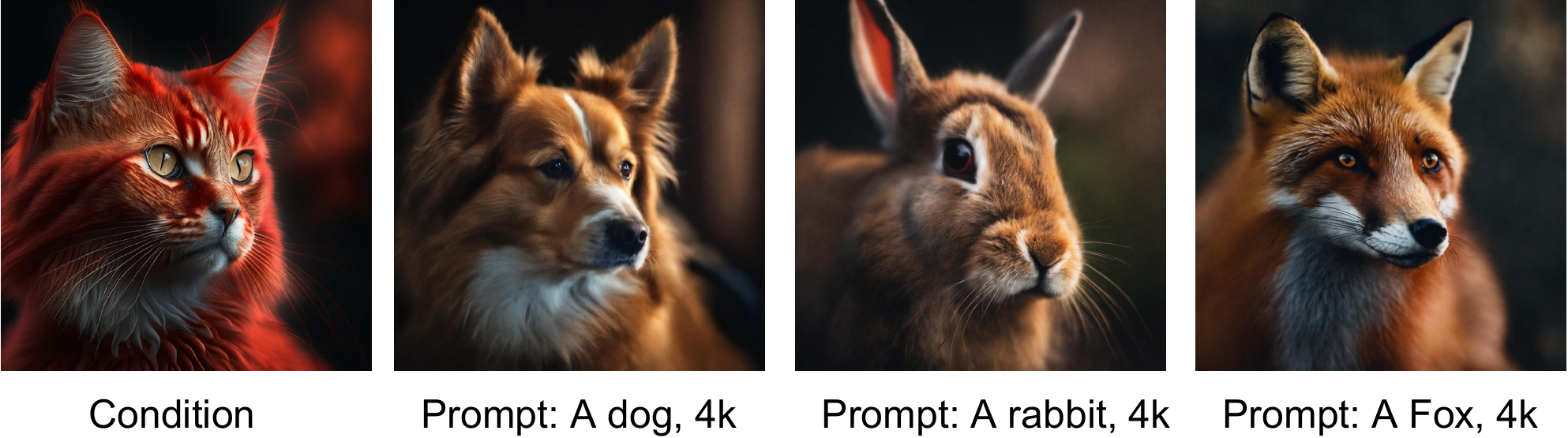}
} \\
\subfloat[Canny Edges to Image]{
    \includegraphics[width=\textwidth]{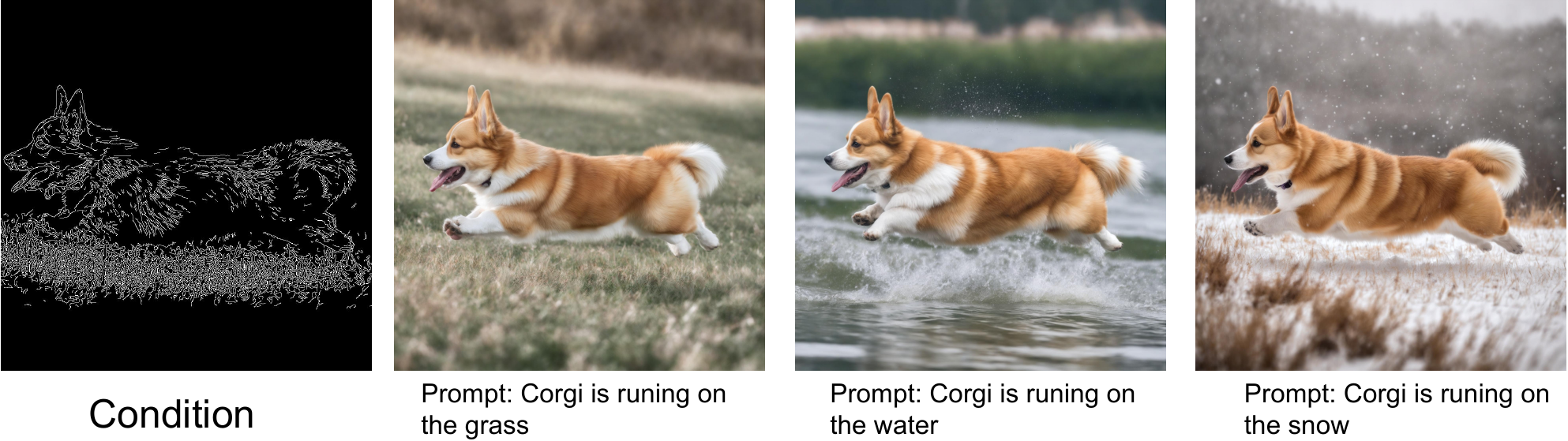}
} \\

\caption{Results of FAM Diffusion combining with ControlNet~\cite{zhang2023adding}. All images are generated at 2× (2048 × 2048).Best viewed when zoomed in.}
\label{fig:our_controlnet}
\end{figure*}

Fam Diffusion enables seamless integration with various LDM-based applications, such as ControlNet~\cite{zhang2023adding}. As shown in Figure~\ref{fig:our_controlnet}, Fam Diffusion combined with ControlNet~\cite{zhang2023adding} achieves controllable high-resolution generation, with examples showcasing the use of images and canny edges as conditions.

\begin{table*}
    \centering
    \begin{tabular}{l|c|ccccc}
    \toprule
        Method & Resolution Scale Factor & $\text{FID}_{r}$ $\downarrow$ & $\text{KID}_{r}$ $\downarrow$ &  $\text{FID}_{c}$ $\downarrow$ & $\text{KID}_{c}$ $\downarrow$ & CLIP Score $\uparrow$ \\
        \hline
         SD 1.5 & \multirow{6}{*}{$2\times2$} & 75.36 & 0.0122 & 43.99 & 0.0103 & 30.35\\
          SD 1.5 + \modelname &  & {65.07} & {0.0087} & {34.06} & {0.0082} & {30.92}\\
         \cline{1-1}
         SD 2.1 & & 86.62 & 0.0163 & 53.67 & 0.0137 &29.66\\
         SD 2.1 + \modelname &  &  {64.77}&  {0.0084}& {38.18} & {0.0091} & {31.13}\\
         \cline{1-1}
         SDXL&  &59.47 & {0.0067} &50.54 & 0.0136 & 30.6\\
         SDXL+ \modelname & &{58.91} & 0.0072 & {{33.96}}  & {{0.0080}}  & { {32.35}} \\
         \hline
         SD 1.5 & \multirow{6}{*}{$3\times3$} & 106.50 & 0.0251 & 48.92 & 0.0133 & 28.89\\
         SD 1.5 + \modelname &  & {38.19} &  {0.0011} & {43.99} & {0.0082}&{30.44} \\
         \cline{1-1}
         SD 2.1 & &137.05 & 0.0384 & 63.91 &0.01719 &27.81 \\
         SD 2.1 + \modelname&  & {64.8} &{ 0.0089} &{ 40.49 }& {0.0114 }&{31.13} \\
         \cline{1-1}
         SDXL& & 78.41 &0.0136 & 69.40 & 0.0210 & 28.44 \\
         SDXL + \modelname &  & {69.25} & {0.0007} & {36.40} & {0.0100} & {32.25}  \\
         \hline
         SD 1.5 & \multirow{6}{*}{$4\times4$} &  150.84 & 0.0474 & 55.97 & 0.0155 & 27.40\\
         SD 1.5  + \modelname &  & {67.77} &  {0.0086}& {40.21} & {0.0012} & {30.36}\\
         \cline{1-1}
         SD 2.1 & & 177.06 & 0.0645 & 69.43 & 0.019 & 26.36\\
         SD 2.1+ \modelname &  & {66.32} & {0.0085} & {41.37 }& {0.0018} & {31.10}\\
         \cline{1-1}
         SDXL& & {160.10} & {0.0602 }& {74.37} &{ 0.0242} &  {26.70} \\
         SDXL + \modelname & & {58.91 }& {0.0073} & {43.65} & {0.0130} & {32.33}
         \\
         \bottomrule
    \end{tabular}
    \caption{ Comparison of vanilla Stable Diffusion and our ~\modelname. }
    \label{tab:ours_plus_sdvar-supp}
\end{table*}
\begin{table*}
    \centering
    \begin{tabular}{l|c|ccccc}
    \toprule
         Method & Scaling Factor  & FID$\downarrow$ & KID$\downarrow$  & FID$_{c}\downarrow$ & KID$_{c} \downarrow$ & CLIP $\uparrow$ \\ \hline
         \cline{3-7}
         DemoFusion \cite{demofusion} & \multirow{7}{*}{$2\times4$} &81.69 & 0.0112 & 54.48 & 0.0165  & 29.3   \\
         AccDiffusion \cite{accdiffusion}&  & 70.42  &0.0119 &  55.73 &0.0205 & 29.0  \\
         FouriScale*~\cite{huang2024fouriscale}&  & 71.86& 0.0302 &  63.28&0.0322  & 25.8  \\
         HiDiffusion \cite{hidiffusion}&  & 118.56& 0.038 & 65.46 & 0.021 & 26.3  \\
         SDXL~\cite{sdxl} &  &80.62 &0.0236 &67.46&0.0302& 25.5\\
         SDXL~\cite{sdxl} + \modelname  &  &\textbf{63.48} &\textbf{0.0090} & {\textbf{41.44}}  & {\textbf{0.0115}}  & { \textbf{30.6}} \\         
         
    \bottomrule
    \end{tabular}
    \caption{System-level comparisons with SDXL. * indicates inference with FreeU~\cite{si2024freeu}}
    \label{tab:comp_to_sota-24}
\end{table*}

\end{document}